\documentclass{article} 
\usepackage{iclr2027_conference,times}

\usepackage[utf8]{inputenc} 
\usepackage[T1]{fontenc}    
\usepackage{hyperref}       
\usepackage{url}            
\usepackage{booktabs}       
\usepackage{amsfonts}       
\usepackage{nicefrac}       
\usepackage{listings}       
\usepackage{microtype}      
\usepackage{xcolor}         
\usepackage{multirow}
\usepackage{tikz}
\usetikzlibrary{patterns}
\usepackage{graphicx}
\graphicspath{{figures/}}
\usepackage{wrapfig}
\usepackage{amssymb}
\usepackage{amsmath}
\usepackage{subfig}
\usepackage{listings}
\usepackage{float}
\usepackage{array}
\usepackage{marvosym}       
\usepackage{fontawesome5}   
\lstdefinelanguage{JavaScript}{
  morekeywords={function, let, const, var, return, if, else, for, while, new,
                typeof, null, true, false, this, break, continue, do, switch,
                case, default, throw, try, catch, class, of, in},
  sensitive=true,
  morecomment=[l]{//}, morecomment=[s]{/*}{*/}, morestring=[b]',
}
\definecolor{jsKeyword}{HTML}{0000C0}
\definecolor{jsString} {HTML}{A31515}
\definecolor{jsComment}{HTML}{2E7D32}
\definecolor{jsNumber} {HTML}{098658}
\definecolor{jsBuiltin}{HTML}{795E26}
\lstdefinestyle{jsfig}{
  emph={createCanvas, noStroke, background, fill, ellipse, rect, keyIsDown,
        width, height, setup, draw, resetGame, getGameState, mulberry32},
  emphstyle=\color{jsBuiltin},
  literate={0}{{\color{jsNumber}0}}1 {1}{{\color{jsNumber}1}}1
           {2}{{\color{jsNumber}2}}1 {3}{{\color{jsNumber}3}}1
           {4}{{\color{jsNumber}4}}1 {5}{{\color{jsNumber}5}}1
           {6}{{\color{jsNumber}6}}1 {7}{{\color{jsNumber}7}}1
           {8}{{\color{jsNumber}8}}1 {9}{{\color{jsNumber}9}}1,
}
\definecolor{jsonKey}{HTML}{6F42C1}
\definecolor{jsonVal}{HTML}{116329}
\definecolor{jsonPunct}{HTML}{57606A}
\lstdefinestyle{jsonfig}{
  language={},
  literate=,
  emph={default8, mouse2d, name, held, press, type, channels},
  emphstyle=\color{jsonKey},
  emph={[2]NOOP, LEFT, LEFT_D, box, pointer_x, pointer_y},
  emphstyle={[2]\color{jsonVal}},
  morekeywords={null, true, false},
  keywordstyle=\bfseries\color{jsKeyword},
  morecomment=[s]{/*}{*/},
  commentstyle=\itshape\color{jsonPunct},
}
\lstdefinestyle{prompt}{
  language={}, basicstyle=\ttfamily\scriptsize,
  frame=none, columns=fullflexible, keepspaces=true,
  breaklines=true, breakatwhitespace=true,
  xleftmargin=0pt, xrightmargin=0pt, aboveskip=8pt, belowskip=8pt,
  lineskip=1.2pt, captionpos=t,
}
\usepackage{amsmath,amsfonts,bm}

\def\eqref#1{equation~\ref{#1}}

\def\1{\bm{1}}

\def\vs{{\bm{s}}}

\DeclareMathAlphabet{\mathsfit}{\encodingdefault}{\sfdefault}{m}{sl}
\SetMathAlphabet{\mathsfit}{bold}{\encodingdefault}{\sfdefault}{bx}{n}

\RequirePackage{hyperref}
\definecolor{darkblue}{rgb}{0, 0, 0.5}
\hypersetup{colorlinks=true, citecolor=darkblue, linkcolor=darkblue, urlcolor=darkblue}

\title{PlayTrain: An Efficient Reinforcement Learning Framework for LLM-Generated Adaptable JavaScript Games}

\author{Ryan Truong$^{1}$ ~ Lance Ying$^{1,2}$ ~ Samuel J.~Gershman$^{1,3}$ ~ Kazuki Irie$^{4}$\\
  $^1$Harvard University, Cambridge, MA, USA \\
  $^2$MIT, Department of Brain and Cognitive Sciences, Cambridge, MA, USA \\
  $^3$Kempner Institute for the Study of Natural and Artificial Intelligence, Cambridge, MA, USA \\
  $^4$Yale University, Department of Computer Science and Wu Tsai Institute, New Haven, CT, USA \bigskip \\  {\normalfont\small \faEnvelope:
  \texttt{truongtruong@fas.harvard.edu} , 
  \texttt{kazuki.irie@yale.edu}} \\
\makebox[\textwidth][l]{\normalfont\small \faGithub: \url{https://github.com/heyodog0/playtrain}} \\[2pt]
\makebox[\textwidth][l]{\normalfont\small \Mundus: \url{https://playtrain.org}}
}

\iclrfinalcopy
\begin{document}

\maketitle
\lhead{Preprint.}

\begin{abstract}
While many video-game environments (VGEs) have played crucial roles in advancing reinforcement learning (RL), developing novel VGEs or modifying existing ones to support new features, has been a laborious process requiring extensive hand-coding.
Here we present PlayTrain, an RL framework that combines the abilities of large language models (LLMs) to robustly generate JavaScript (JS) games from a minimal human prompt, and an efficient pipeline that can run any JS game in a standard `gym' environment. Not only are recent LLMs particularly good at writing JS code, but the JS format also allows users to easily \textit{play} generated VGEs, while PlayTrain enables us to \textit{train} RL agents on the exact same games.
We demonstrate multiple use cases of PlayTrain, including cloning well-known Atari and ProcGen games in simple JS, where PlayTrain trains pixel-based agents end-to-end at over 1M agent-decisions per second on a single GPU node; and creating modified versions thereof (e.g., that support novel test sets, procedural generation logics, or game dynamics).
Through PlayTrain, we reimagine RL VGE development: all we need is a single JS file, generated and modified through an LLM. We discuss promising future RL research directions that PlayTrain unlocks.
\end{abstract}

\section{Introduction}
Video games have long played a critical role in the development of reinforcement learning (RL) algorithms. From early milestones in classic Atari games \citep{mnih2015dqn, bellemare2013ale} to superhuman mastery in complex strategies like \mbox{StarCraft II} \citep{vinyals2019starcraft} and \mbox{Dota 2} \citep{berner2019dota}, games have driven major advancements in
artificial intelligence (AI) systems. Going beyond existing games, RL researchers have also designed novel game environments, allowing them to devise custom tasks to test specific abilities of an intelligent system \citep{beattie2016deepmind, cobbe2020procgen, kuttler2020nethack, matthews2024craftax}---posing new challenges of game engineering for RL.\looseness=-1

What, then, defines an ideal game environment and training framework for  RL research?
\textit{Efficiency} is paramount, as environment interactions typically represent the primary speed bottleneck in the training pipeline; \textit{conceptual complexity} is equally critical, because the environment defines the boundaries of what can be learned---agents trained on uninteresting environments are inevitably limited.
In addition to these two classic aspects, there are two other crucial, yet often overlooked, properties: \textit{adaptability}---that is, how easily one can modify the environment to accommodate new ideas; and \textit{playability}---how easily a human can interact with and test the environment to ensure it accurately reflects the intended research goals or to compare RL agents against humans.

\begin{figure}[t]
\centering
\includegraphics[width=.90\linewidth]{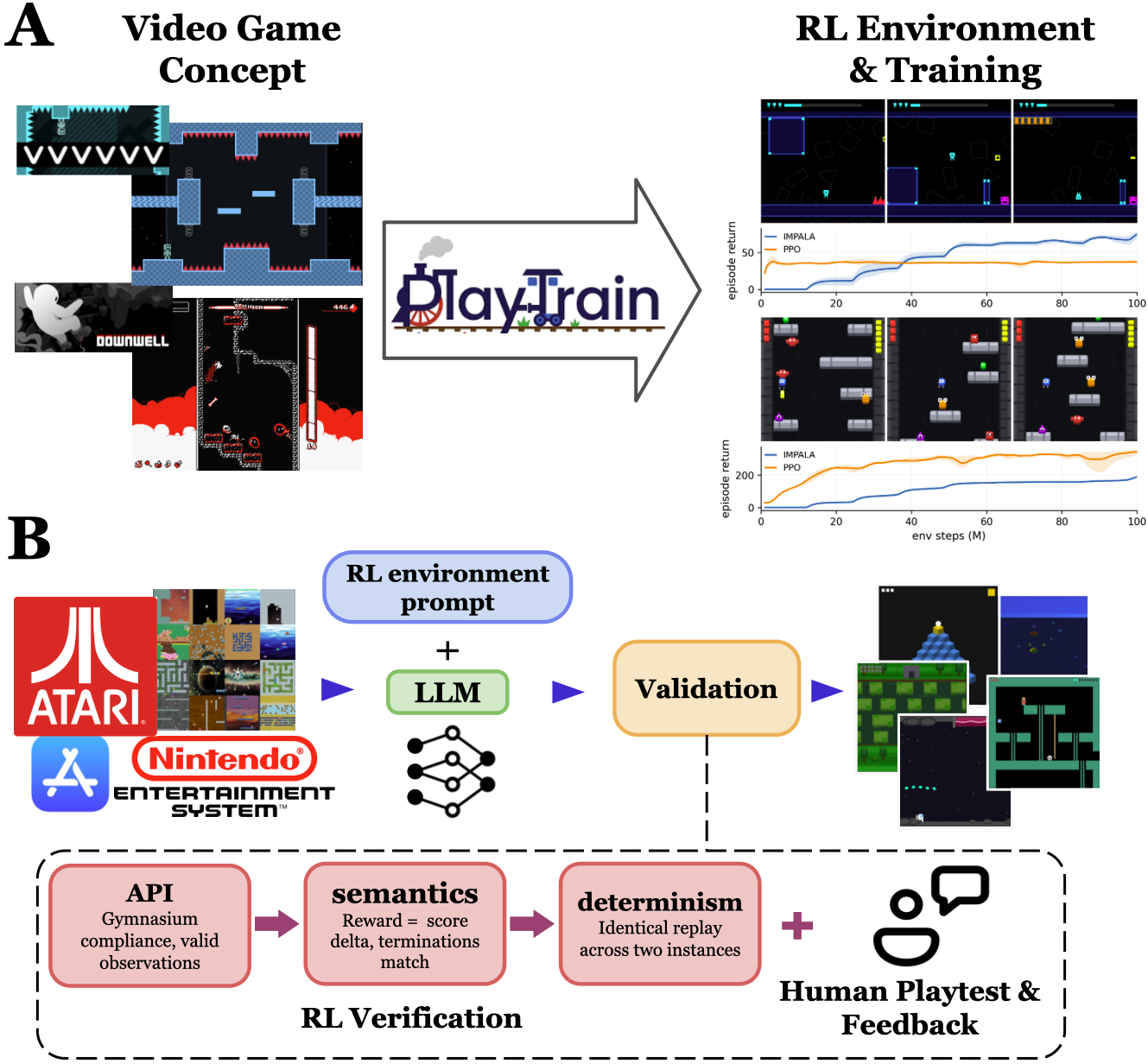}
\caption{\textbf{A:} The concept of PlayTrain: Through LLM-generated JS code, PlayTrain can transform games such as ``DownWell'' and ``VVVVVV'' into an  RL environment and train agents in minutes.
\textbf{B:} The generation pipeline. From a short specification of a video game, an LLM generates a single self-contained JavaScript file against a fixed RL-environment prompt-file, creating an environment that is both immediately playable and ready for baseline RL agent training without any additional code.\looseness=-1}
\label{fig:generation}
\end{figure}

Here we describe PlayTrain (Figure \ref{fig:generation}), an RL framework based on JavaScript (JS) that integrates these four elements.
In fact, JS itself offers two of these properties---complexity and playability---by design, as a popular high-level language for game development.
JS allows us to improve over the classic RL environments on the \textit{complexity} axis, as it offers a larger space of representable mechanics and variations for game design, surpassing the bounded complexity of classic games, e.g., the fixed ROM set from ALE or ProcGen's parametrized generators.
\textit{Playability} is also given by construction, because every game or environment built in JS can run on the browser and is shareable through a single link or file---no extra package installation is required. Generated games can be effortlessly played and tested, not only by the developer, but also by any testers.\looseness=-1

The core challenge for JS is \textit{efficiency}, which is the main reason why JS has been considered unsuitable for building RL environments: browser-based processing is slow and the resulting slow environmental interactions bottleneck any practical RL training pipelines in terms of steps-per-second (SPS) throughput.
Here we challenge this common belief by proposing a framework that overcomes this bottleneck and achieves environment efficiency that exceeds well-established benchmarks.


With the efficiency bottleneck resolved, the JS-based framework offers an excellent synergy with LLMs to unlock unparalleled \textit{adaptability} for RL environment development, as it is a higher-level language (e.g., compared with C++) that's easier for both humans and the current best LLMs to read and modify to develop games.
Historically, creation of variations of RL environments has been bounded to a handful of ideas, fixed in advance---most often level layout and random seeds \citep{cobbe2020procgen, cobbe2019coinrun}, and rules composed inside a domain-specific language \citep{schaul2013vgdl, bamford2021griddly}. 
While such features have played important roles in RL research to go beyond \textit{testing on the training data}, those variations have been still very limited for evaluating broader generalization of RL agents \citep[see for example][]{kirk2023survey, ShanahanM22}. Yet, development of more flexible RL environments has been challenging due to laborious engineering efforts required for implementing new environments and testing them.

Through PlayTrain, we aim to substantially facilitate this development process and accelerate the entire RL research pipeline, from the conception of a game environment to RL training.
PlayTrain also enables users to flexibly edit environments or create variations thereof---e.g., to construct novel test environments---by modifying a game's visuals, physical parameters, or mechanics, or even by transplanting the dynamics of one game into another, typically via a single natural-language prompt; the only limitation is our ability to describe, reflecting our motto:
``\textit{What I can describe, I can create.}''\looseness=-1




\section{Methods}

PlayTrain consists of the following components: (1) an effective LLM-prompting pipeline that leverages LLMs' proficiency in generating and modifying JS game environments (Sec.~\ref{sec:generation}), and (2) a novel JS-to-gym backend that achieves high efficiency in running JS games for training of RL agents (Sec.~\ref{sec:backend}).
As a result, PlayTrain enables an unprecedentedly efficient transition from the conception of a game environment to its actual implementation, and the training and testing of RL agents and algorithms on the generated environment (see Figure \ref{fig:generation}A for illustration).\looseness=-1


\subsection{Generating and developing JS environments using an LLM}
\label{sec:generation}

An overview of PlayTrain's environment generation pipeline is illustrated in Figure \ref{fig:generation}B.
It starts with a short specification stating the name of the environment/game, 
its core mechanics, and possibly online references.
This information is fed into an LLM (Google's Gemini 3.1 pro) with a Markdown prompt file that specifies it to export components of an RL lifecycle such as \texttt{setup}, \texttt{draw}, \texttt{resetGame(seed)}, and \texttt{getGameState()}. The prompt can be found in Appendix~\ref{app:prompt}. 
The output is a single self-contained JavaScript file. 

After the LLM-based generation,
the next step in the PlayTrain pipeline is an automated, game-agnostic validation pass. 
A script runs the environment to confirm Gymnasium API \citep{towers2024gymnasium} compliance and ensure the environments generate proper observations. 
After, the script replays a random action sequence in two separate instances, requiring the two observation streams to be identical. 
Crucially, the check is only a single script and involves no human inspection. While all of the environments we generated in our experiments (Sec.~\ref{sec:experiments}) passed these requirements one-shot, the validation pass ensures that the generated games and their variants are RL-compatible. 


To obtain the final game environment file, the LLM-based generation may not always be one-shot: play-testing may reveal discrepancies between the generated game and the intended game design.
Luckily, play-testing a JS game itself is a straightforward process (essentially, opening a link on a browser) and the game refinement process is easy because LLMs are very good at editing and correcting JS code
through simple natural-language feedback. \looseness=-1

Producing variants (e.g., a novel test environment) of an already-created game is also just as simple as the game refinement process.
This typically only requires a single prompt, making the process of creating variations scalable, unlike with prior, classic RL environment code. We provide illustrative examples in the experimental section.
Full generation details such as the catalog schema, prompt template, and model configuration are provided in Appendix \ref{app:prompt}.




\subsection{Efficient backend to support JS environments for RL}
\label{sec:backend}

\begin{figure}[t]
\centering
\includegraphics[width=1\linewidth]{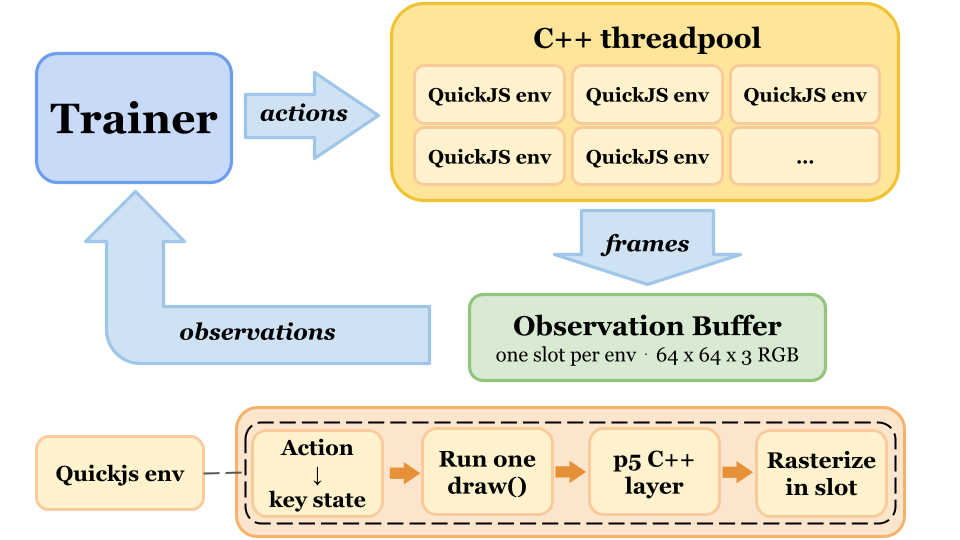}
\caption{PlayTrain training schematic. A trainer sends actions to multiple environment instances hosted in an in-process C++ threadpool. Each instance is an embedded QuickJS engine running the game file behind our optimized C++ p5 library and custom Rust rasterizer. Each  instance writes its frames directly into its own slot of the shared observation buffer ($64{\times}64{\times}3$ RGB), which the trainer reads as observations.
In each instance: per step, the environment receives an action and transforms it into a keyboard state and then runs a single \texttt{draw()}. Our C++ p5 API intercepts those draw calls, and our rasterizer executes them at observation resolution directly into that environment's slot.}
\label{fig:backend}
\end{figure}

Another unique component of PlayTrain is our novel backend to efficiently process JS environments for RL training loops.
The key for efficiency is running JS browser games \textit{without} a browser.
In fact, JS itself is not what limits throughput in an RL setting: the limit comes from its dependency on the browser to render frames and receive/inject inputs. 
A browser itself carries a great deal of 
``slow'' machinery and protocol that RL agents do not need. PlayTrain removes these unnecessary components and only keep what's needed for training RL agents, which reduces the whole pipeline to only two things:
an \textit{input channel} and a \textit{renderer}. 





\textbf{Input Channel and Renderer.} 
The input channel is simple: an agent writes actions into the game's key states which are passed to be read by the game (Appendix~\ref{app:pt_act_obs}).
The renderer is more complex---and it is precisely where PlayTrain improves over existing solutions: 
everything a game draws goes through one library, p5.js \citep{p5js}, which provides commands like \texttt{ellipse(x, y, w, h)} to draw an ellipse of a given size/position, and \texttt{fill(r, g, b)} to set the color that everything drawn after will use.
A PlayTrain environment collects those commands inside a function named \texttt{draw()}, and running it once issues every command needed for a single frame. Turning those \textit{draw commands} into pixels takes three steps: (1) the game's code runs, (2) the draw calls become shapes with positions and colors, and (3) those shapes render into pixels.\looseness=-1

We design PlayTrain to handle all three steps optimized for speed.
In PlayTrain, the Javascript code runs on QuickJS \citep{quickjsng, bellard2024quickjs}, a small engine that ships as a C library, which we compile directly into each PlayTrain environment.
(more details presented in Appendix~\ref{app:backend}). 
Alongside it we compile a p5 library we rewrote in C++, 
defining a subset of the same function names the original p5.js library uses. 
Because both are compiled into the \textit{same program}, QuickJS can call our p5 C++ functions directly, enabling that a game's call to \texttt{ellipse} or \texttt{fill} run our code and lets the generated environment file run \textit{as written}. 

For the third step---rendering shapes into actual pixels---we wrote a custom rasterizer in Rust \citep{matsakis2014rust} that is compiled in tandem with our C++ p5 library. 
It therefore shares memory with the draw commands that library intercepts and writes pixels into the trainer's observation buffer \textit{directly} with no browser in the loop.

\begin{figure}[t]
\centering
\includegraphics[width=\linewidth]{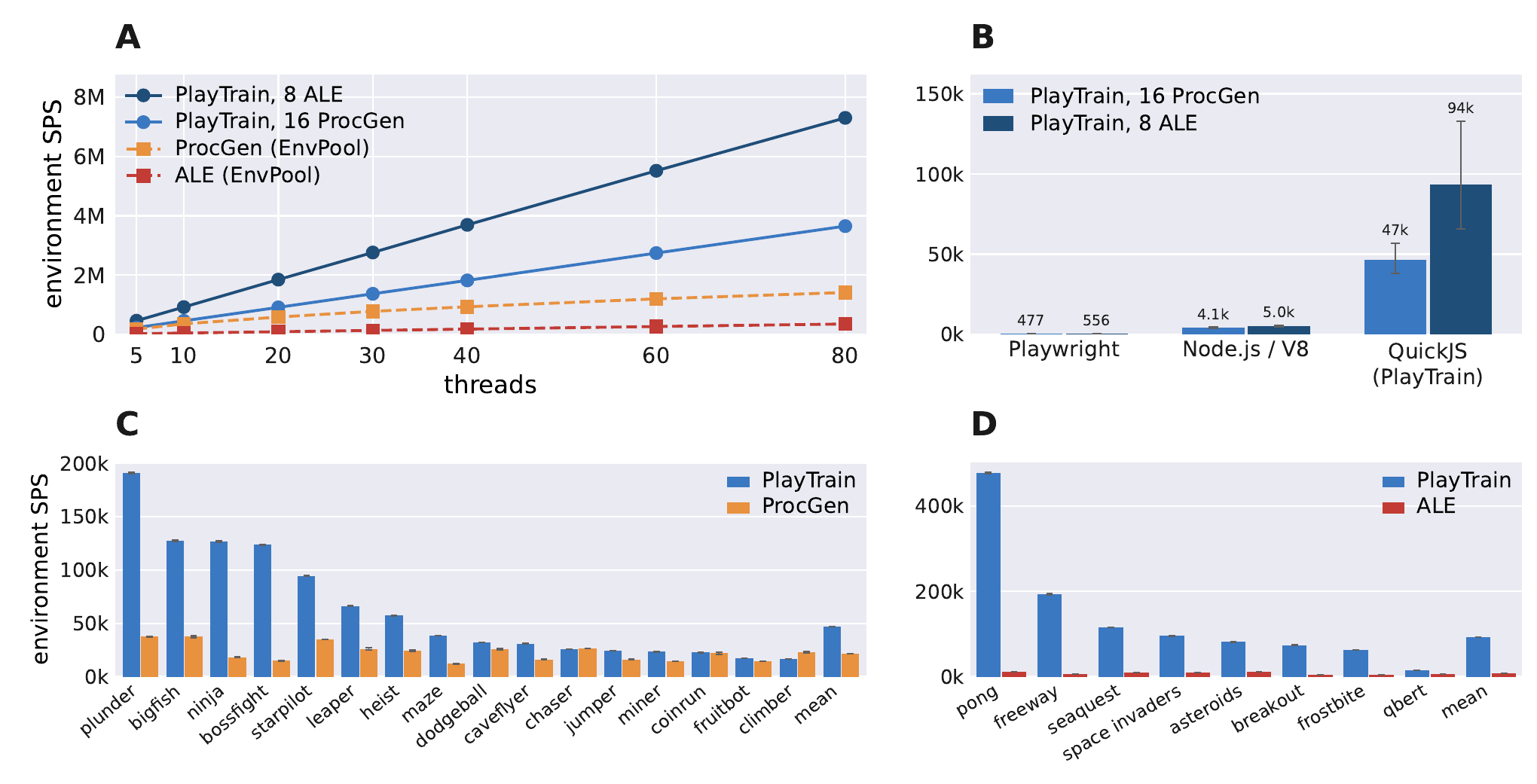}
\caption{\textbf{(A)} Environment step speed (without a  trainer) as a function of the number of threads. 
Each line is a geometric mean over one suite, and the baselines are the original C\texttt{++} environments in EnvPool.
\textbf{(B)} Environment steps per second for the same games under three backends: a headless browser driven through the Playwright API, a standalone Node/V8 engine, and PlayTrain's QuickJS with the native rasterizer, split into the 16 ProcGen and 8 ALE replicas. Bars are geometric means over games; error bars are $\pm1$ s.e. across games (log space, asymmetric).
\textbf{(C, D)} Per-core environment throughput (without a trainer) for each game and the geometric mean.\looseness=-1}
\label{fig:env_efficiency}
\end{figure}

\textbf{Training.} 
In practice, training typically spawns multiple environments simultaneously for improved efficiency.
That usually means running each environment as its own program \citep{towers2024gymnasium}.
Separate programs, however, cannot see each other's memory. So, in that setup, every observation would have to be \textit{copied} to the trainer, which is a slow operation \citep{petrenko2020samplefactory}.
PlayTrain instead steps its environments on threads inside the \textit{same program} as the trainer, a design adapted from EnvPool \citep{weng2022envpool}.
Threads are independent lines of execution inside a program: many run at once on different cores while they all share the same memory. Each rendered frame is therefore written straight into the buffer the trainer reads from.
As a result, stepping scales almost linearly with the number of available threads (Figure~\ref{fig:env_efficiency}A). 

\textbf{Interface and Reproducibility.} An agent's action space and observations are defined by the PlayTrain backend.
More precisely, a game reads keyboard inputs as it would in a browser.
This design is intentional because it enables every game to be \textit{agnostic} to any trainer and would otherwise require mapping a custom action space and observation size onto the trainer for every game. 
By default, PlayTrain's action space are 8 discrete actions with observations set to 64$\times$64 RGB, yet the action space and observations are all flexible and modifiable (Appendix~\ref{app:pt_act_obs}). 
Runs are also fully reproducible: every game is stepped with a single \texttt{draw()} call, every game-logical source of non-determinism is fixed with a seed generator \citep{mulberry32}, and our rasterizer's inner-workings generate identical pixels across machines. In effect, the game files that models train are equal to the files humans play on as well. 
(Appendix~\ref{app:backend}).

\section{Experiments}
\label{sec:experiments}

Here we present several illustrative experiments that demonstrate the capabilities and efficiency of PlayTrain, including speed comparison with classic benchmarks (Sec.~\ref{sec:efficiency}), examples of various environments it can generate (Sec.~\ref{sec:variants}), and an example collecting human playing data (Sec.~\ref{sec:human}).

\subsection{Efficiency of Environments and Training}
\label{sec:efficiency}

We first demonstrate that PlayTrain is an efficient framework for RL training by comparing its speed with classic benchmarks.
For that, we use PlayTrain (Sec.~\ref{sec:generation}) to generate eight clones of the classic Arcade Learning Environment games \citep{bellemare2013ale} and 16 clones of ProcGen \citep{cobbe2020procgen} environments, for the total of 24 games, with the goal of measuring the speed of each replica against the original environment.
While certain details (e.g., the exact game visual) may be different,
PlayTrain can create high-quality clones of existing games in JS---reproducing the core game mechanics and dynamics and emulating the spirit of the originals (see screenshots in Figure \ref{fig:learning}A).
We release all the generated games in our repository so that their fidelity can be verified.
\begin{table}[t]
\begin{minipage}[c]{0.40\linewidth}
\caption{Single-node training throughput, agent-steps/s at frame skip 1 and
$64{\times}64{\times}3$ RGB on one node with four H100s and 92 CPU cores,
geometric mean over the games in each row. The original C\texttt{++}
environments in (b) are hosted in EnvPool. IMPALA rows marked with $\dagger$ are
\textit{double-buffered} and PPO rows are single-buffered. (b) reports single-buffered speed using IMPALA with the
Nature-CNN encoder, so we can compare against EnvPool which has no
double-buffered option. Single-buffered PlayTrain in (b) is naturally slower than its double-buffered counterpart of (a).\looseness=-1}
\label{tab:train-throughput}
\end{minipage}\hfill
\begin{minipage}[c]{0.57\linewidth}
\centering
\vspace{-5mm}
{\footnotesize
\setlength{\tabcolsep}{0.4em}
{\itshape (a) PlayTrain environments, double-buffered}\\[2pt]
\begin{tabular}{@{}lllr@{}}
\toprule
Trainer & Encoder & Envs & Agent-steps/s \\
\midrule
IMPALA$^\dagger$ & Nature-CNN & all 24     & \textbf{1.07M} \\
 & IMPALA-CNN & all 24     & 0.35M \\
PPO    & Nature-CNN & all 24     & 185k \\
    & IMPALA-CNN & all 24     & 68k \\
\midrule
IMPALA$^\dagger$ & Nature-CNN & 16 ProcGen & 1.06M \\
 & Nature-CNN & 8 ALE      & 1.09M \\
\bottomrule
\end{tabular}

\vspace{0.7em}

{\itshape (b) PlayTrain clones vs. originals, single-buffered}\\[2pt]
\begin{tabular}{@{}lllrr@{}}
\toprule
Trainer & Encoder & Envs & Original & PlayTrain \\
\midrule
IMPALA & Nature-CNN & 16 ProcGen & 372k & \textbf{838k} \\
 & Nature-CNN & 8 ALE      & 175k & \textbf{1{,}018k} \\
\bottomrule
\end{tabular}}
\end{minipage}
\vspace{-5mm}
\end{table}

\textbf{Environment efficiency.} 
We first measure the pure environmental speed without training an agent, that is, the number of environmental steps a single PlayTrain replica produces per second on \textit{a single core}.
Figure~\ref{fig:env_efficiency}C and D shows the results: PlayTrain's JS replicas outpace ALE on all eight shared games and ProcGen's hand-written C++ on fourteen of the sixteen, with speedups of 12.62$\times$ and 2.18$\times$, respectively, on geometric average over suites.
The remaining two ProcGen clones are not faster than the originals but the speed is still respectable.
Figure~\ref{fig:env_efficiency}B shows how much the backend matters: the same games run in a headless browser through the Playwright API, on a standalone Node/V8 engine, and on QuickJS (PlayTrain) compiled into the environment itself. 
QuickJS steps them 13.4$\times$ faster than Node/V8 and 117$\times$ faster than the browser (Appendix~\ref{app:playwright_node_qjs}).

The situation becomes even more favorable for PlayTrain in the more realistic \textit{multi-thread} multi-environment setting with a trainer attached.
Table \ref{tab:train-throughput} shows the results.
Here, all PlayTrain clones are faster than the original ProcGen games, by a factor of 2.25 $\times$ on average and is also 5.81$\times$ faster than ALE on all eight.
The ALE ratio is 5.81$\times$ here rather than 12.62$\times$ measured per core because the PlayTrain training runs' speeds are restricted by the trainer.
Without any trainer in the loop, the same environments reach 20.80$\times$ ALE and 2.58$\times$ ProcGen at eighty threads (Figure~\ref{fig:env_efficiency}A).
This is because PlayTrain environments scale linearly while Envpool's ProcGen flatten (Appendix~\ref{app:bench}), which is remarkable.  
EnvPool's ALE doesn't flatten, but our 8 game suite retains the 20$\times$ ratio. 

\textbf{Training efficiency.}
Now we evaluate speed of the end-to-end RL training process.
Following a common standard, we measure the speed for two classic vision encoders: Nature-CNN \citep{mnih2015dqn} and IMPALA-CNN \citep{espeholt2018impala}; and two classic algorithms, PPO \citep{schulman2017ppo} and IMPALA. Other training and policy hyper-parameters can be found in Appendix~\ref{app:trainer}.

Table~\ref{tab:train-throughput} shows the results:
PlayTrain trains the 24-game suite in an average of 1.07~M agent-steps per second under IMPALA with the Nature-CNN encoder---23 out of the 24 games surpass 1M steps per second. \texttt{climber} is the only one that sits below, training at 881,299 steps per second.
With the IMPALA-CNN encoder, the speed of 0.35~M agent-steps per second is achieved with the node's four GPUs split two to the learner and two to inference.
Under this encoder specifically, the learner becomes the sole bottleneck so throughput barely varies by game.
A large part of this speed comes from a method called \textit{double buffering}: 
one group of environments steps while agent inference runs on the other group of environments so that stepping and inference can be processed in parallel (Appendix~\ref{app:trainer}). 
This is also why the PlayTrain numbers in Table~\ref{tab:train-throughput}(b), which run single-buffered, sits below their counterparts in (a).

Figure~\ref{fig:learning}C shows training curves for eight representative games.
The rest of the games' learning curves are presented in Appendix~\ref{app:suite}.
We use one common trainer configuration for all the games without any game specific tuning (see Table~\ref{tab:hyperparams} in the appendix).
Every episode draws a new seed, so training runs on the unbounded level distribution rather than a fixed set of levels.
In Table~\ref{tab:eval}, we report the mean and 95\% confidence interval (CI) over three seeds and evaluate the final checkpoints on 8 held-out seeds, and compare against a random policy baseline. Results are in Appendix~\ref{app:suite}. 

\subsection{Generating variants and new games}
\label{sec:variants}

\begin{figure}[t]
\centering
\includegraphics[width=1\linewidth]{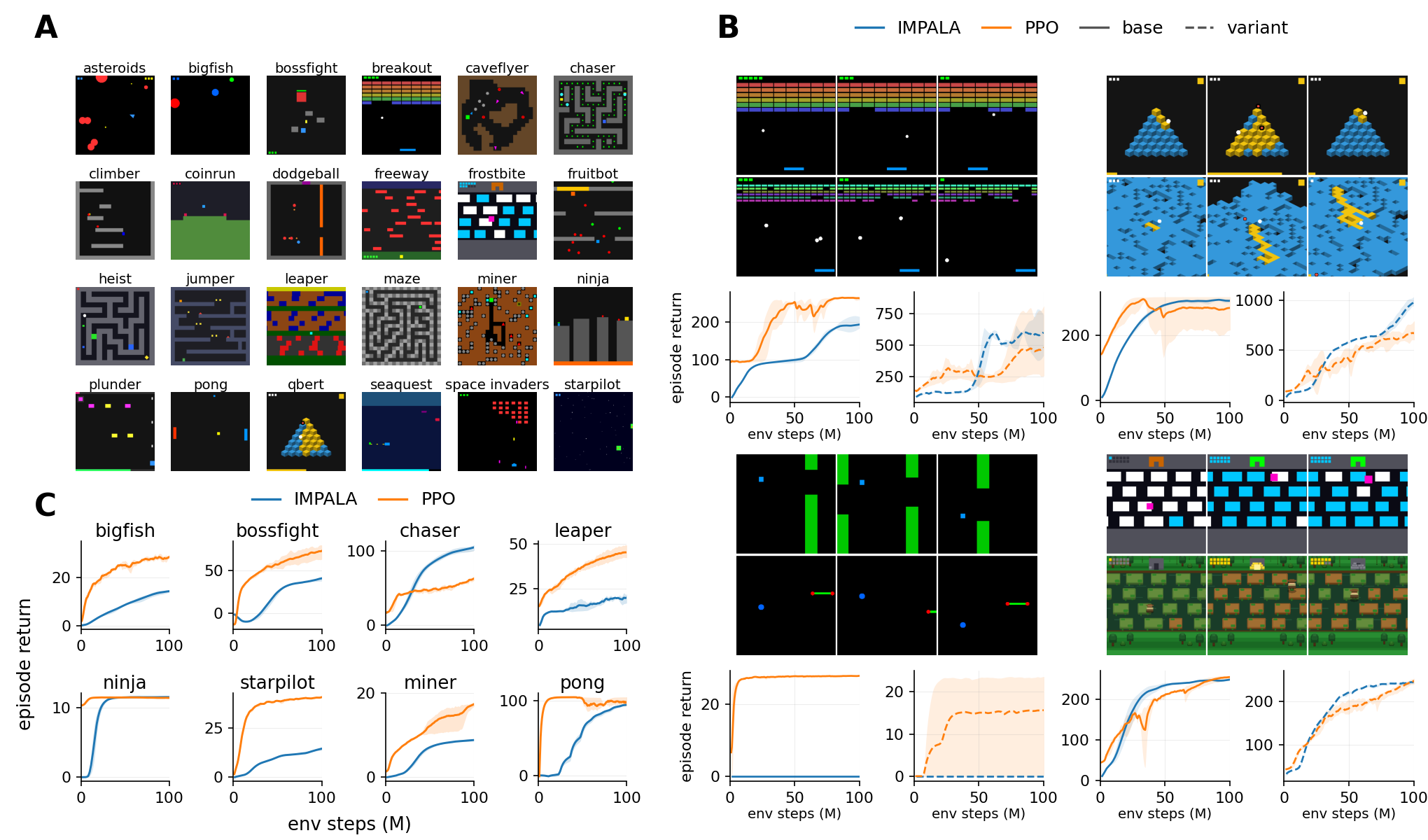}
\caption{Illustrations of PlayTrain-generated environments and training curves. \textbf{(A)} 24 game examples, cloning Atari and ProcGen games. \textbf{(B)} Four example pairs of base game vs.~new variant, with training curves; left to right, top to bottom: \texttt{breakout} / \texttt{breakout.multiball}, \texttt{qbert} / \texttt{qbert.bigmap}, \texttt{flappy\_bird} / \texttt{flappy\_bird.hoop}, \texttt{frostbite} / \texttt{frostbite.jungle} (generation details in Appendix~\ref{app:prompt}). \textbf{(C)} Training curves for PPO and IMPALA agents on eight representative games; both configurations use the IMPALA-CNN encoder; 3 training seeds are used in all cases (all 24 games in Appendix~\ref{app:suite}).  
}
\label{fig:learning}
\end{figure}

Here we demonstrate how seamlessly PlayTrain can generate (1) a wide variations of existing environments, 
and (2) novel RL environments derived from pre-existing game concepts. 
Figure~\ref{fig:learning}B shows four examples displaying the newly generated variations and the original games, side-by-side.
Three of the four examples correspond to variations of the ALE games already mentioned above:
\texttt{breakout}, \texttt{Qbert}, and \texttt{frostbite} (examples for (1));
the last example
\texttt{flappy\_bird} is a clone of a popular mobile game, as an example for (2).

Through these examples, we illustrate three representative ways of creating variations of environments, which we refer to as: parametric, structural, and visual/thematic variants:


\textbf{Parametric} variants are created by modifying values of certain variables that play a key role in the game, such as gravity, NPC speeds, or ranges of certain variables used in procedural generation (e.g. number of entities).
For example, using PlayTrain, we generated \texttt{breakout.multiball} which is a variant of \texttt{breakout} where the bricks shrink from 8 columns of $46{\times}16$px to 16 at $21{\times}8$px, three balls are in play at once, and a lost ball is permanent rather than respawning.

\textbf{Structural} variants change the structure of the world itself,
such as new map layouts requiring novel strategies or larger maps that stress exploration.
\texttt{qbert.bigmap} replaces the static pyramid of the standard  \texttt{qbert} with a flat, far larger map that spans the whole screen and pans with the agent. 
Another example is \texttt{flappy\_bird.hoop}. In the existing \texttt{flappy\_bird}, the agent flies through gaps between obstacles. 
In the variant, it instead has to fall through a hoop, like a ball scoring a basket. This variant intentionally retains the same action space as the original.

\textbf{Visual/thematic} variants change how a game looks while keeping its game mechanics fixed.
Here, \texttt{frostbite.jungle} is exactly the same game as the classic \texttt{frostbite}, except that its theme is changed from the arctic survival to survival in a jungle, with the corresponding visual modifications.

While we limit ourselves to these few examples due to space limitation, PlayTrain supports many other ways to create variations (e.g., introducing new actions).

Now, instead of modifying existing RL environments to generate their variants, we show that PlayTrain can also help us build novel environments from a game concept alone. 
Here we show two such examples, each generated from a single prompt and refined with a handful of simple natural-language feedback rounds: \texttt{VVVVVV}, a 2010 platformer in which the avatar flips its gravity vertically to collect items and progress, and \texttt{Downwell}, a 2015 action platformer in which an avatar with downward-firing boots descends a well of enemies and gems, stomping and shooting as it collects (Figure~\ref{fig:generation}A).


To illustrate that PlayTrain also facilitates training and evaluation of RL algorithms on the generated games,  we share results for PPO and IMPALA agents trained on both the generated  variants and the original games in Figure~\ref{fig:learning}B. Note again that these are for illustrative purpose; we did not perform any game specific hyper-parameter tuning.



\begin{figure}[t]
\centering
\includegraphics[width=\linewidth]{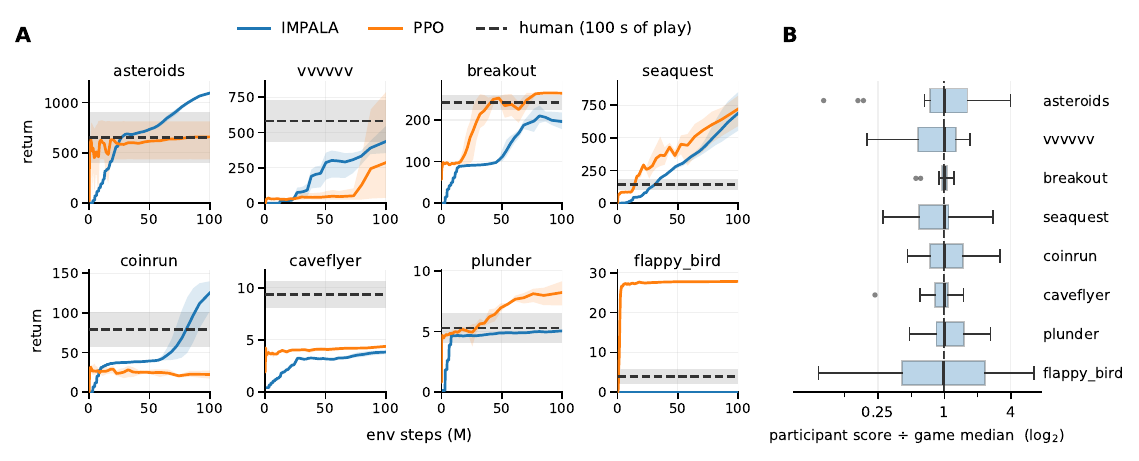}
\caption{\textbf{(A)} Comparing IMPALA and PPO agent training efficiency against 100 seconds of human play. 
Dashed horizontal lines are the human mean over 20 participants and the gray band is the 95\% confidence interval. 
Solid curves are the RL agents' means over three seeds, shaded with min--max bands. 
Both RL agents use the IMPALA-CNN encoder.  
\textbf{(B)} The distribution of the 20 participants' scores on each game. 
Each participant's score is divided by that game's median so that all the games can share the same axis.}
\label{fig:human_wallclock}
\end{figure}

\subsection{Human play}
\label{sec:human}

Here we highlight the human playability strength of PlayTrain and how it facilitates human studies.
As an illustrative example, we collect human performance on some of the PlayTrain-generated games discussed above, and compare to RL agents' performance.
For that, we recruited 20 participants on Prolific (mean age $32.4$, standard deviation $9.0$, range $19$--$54$; 6 women, 13 men, 1 non-binary).
Each played eight games shuffled randomly---\texttt{asteroids}, \texttt{breakout}, \texttt{seaquest}, \texttt{caveflyer}, \texttt{coinrun}, \texttt{plunder}, \texttt{flappy\_bird}, and \texttt{VVVVVV}---for 100 seconds per game, with only instructions about the controls.
PlayTrain allows participants to play the same game the agents train on; here with the episode length capped at 2000 steps and the same seeds for everyone so that they are all evaluated on identical levels.
A participant's score on a game is the mean over their episodes, and the human mean is the mean of those 20 scores. 

Such human experiments allow us to answer interesting questions at the intersection of cognitive science and AI. For example,
how much experience does an agent need to match what a person scored in 100 seconds of play on these specific games?
Figure~\ref{fig:human_wallclock} shows the corresponding results.
On six of the eight games, our PPO or IMPALA trainers reach the human mean, and which trainer succeeds is game-dependent: PPO is the only one to learn \texttt{flappy\_bird},
reaching the human mean at 1~M steps, while IMPALA is the only one to reach it on \texttt{coinrun}, and only after 81~M.
Neither reaches it on \texttt{caveflyer} or \texttt{VVVVVV} (Figure~\ref{fig:human_wallclock}A).

Again, this only represents a simple illustrative example to show how PlayTrain provides a seamless pipeline to allow humans and RL agents to play the exact same game. We leave potentially more complex and deeper human/machine comparison studies that PlayTrain unlocks for the future work.

\begin{table}[t]
\caption{Qualitative properties of RL environment/framework families. 
``GPU-port'' covers Octax \citep{radji2025octax}, and PuzzleJAX
\citep{earle2025puzzlejax}, which port environments and games onto accelerators.
PlayTrain is the only framework that is fast to train on and easy to modify, simultaneously.}

\setlength{\tabcolsep}{0.3em}
\centering
\small
\begin{tabular}{l|cccc}
\toprule
Property & Atari & ProcGen & GPU-port & PlayTrain (ours)\\
\midrule
Complexity & Acceptable & Acceptable & Acceptable & \textbf{Flexible} \\
Efficiency/Speed & Acceptable & High & High & \textbf{High} \\
Adaptability \\
\quad Training variations & Almost None & Procedural & Engine Bound & \textbf{Anything Describable} \\
\quad Test environments & None & Procedural & Engine Bound & \textbf{Anything Describable} \\
\quad Game designs/dynamics & Low & Low & Low &  \textbf{Very High} \\
Human Playability & Medium & Low & High & \textbf{High and Adaptable }\\
\bottomrule
\end{tabular}
\centering
\label{tab:contrast}
\end{table}

\section{Discussion}
\label{sec:discussion}

\textbf{Further Related Work.} In addition to the references cited above, there is further prior work on both developing efficient RL frameworks and creating novel environments to evaluate RL generalization. 
Prior efforts to automate environment creation either generate an environment as a neural world model \citep{bruce2024genie} or have an LLM generate task code for a fixed simulator \citep{faldor2024omniepic, zala2024envgen}; neural world models cannot be inspected or edited though, and fixed simulators are confined by the simulator's engine. 
Other efforts were designed for speed and done so by either porting engines to the GPU \citep{dalton2020cule, shacklett2023madrona} or by batching handmade C++ environments \citep{weng2022envpool}, both of which deliver speeds at the cost of making mechanics harder to modify. 
Appendix \ref{app:related} discusses this prior work in detail, and Table \ref{tab:contrast} provides an overview of PlayTrain's unique features, simultaneously achieving complexity, efficiency, adaptability, and playability.\looseness=-1

\textbf{Scope Limitations.} PlayTrain is limited in the size and complexity of the games it can produce. 
Every game in this paper is a single JS file of a few hundred lines that is generated by an LLM and refined with natural language prompt edits.
One cannot, however, simply prompt PlayTrain for a full-fidelity clone of a modern console game, such as \textit{Legend of Zelda: Breath of the Wild}, since current LLMs cannot reliably generate a game of that size.
Furthermore, our framework focuses on 2D environments and doesn't yet fully support 3D games.
These limitations, however, are not permanent;
LLM capabilities are advancing rapidly \citep{kwa2025horizons, jimenez2024swebench}, so the size and complexity of the games these models can express will continue to improve.

\textbf{Speed Limitations.} PlayTrain's speed declines with the ``work'' a game does per frame, and either drawing or game logic can become the bound.
Drawing calls binds it when a game issues many draw calls: \texttt{qbert.bigmap} reaches 39k SPS against the suite's 0.35M ceiling (Table~\ref{tab:llm-cost}), and \texttt{miner} spends 75\% of its step on 787 drawing commands (Appendix~\ref{app:envcost}).
Game logic, on the otherhand, binds it when a game updates a large amount of state per step, as in \texttt{dodgeball} and \texttt{climber}, whose steps are only 10\% and 18\% drawing, and is why \texttt{climber} is one of the two clones still slower than its original. 

\textbf{Further potential of PlayTrain.}
Beyond the examples shown here, one can use PlayTrain to easily turn many other games previously unsupported for RL into RL environments (similar to VVVVVV and Downwell).
And because the same JS game is playable by a human, an RL agent, or a VLM agent, Playtrain makes performance across these learners directly comparable. 
Moreover, methods whose bottleneck is the number and diversity of training environments rather than the algorithm itself, such as meta-RL \citep{oh2025discovering}, co-evolution \citep{wang2019poet}, and unsupervised environment design \citep{dennis2020paired}, can benefit from PlayTrain's ability to expose new mutation variables or entire new environments through a single LLM prompt. 
Going a step further, PlayTrain's ultimate potential lies in accelerating the process of generating \textit{brand new} game environments, unlocking directions in RL research previously limited by the difficulty of environment development.
We further discuss such directions in Appendix \ref{app:furtherdisc}.

\section{Conclusion}
\label{sec:conclusio}

With PlayTrain, we reimagine RL research by unifying efficiency, adaptability, complexity, and playability.
PlayTrain generates adaptable JS environments with a large language model and provides an efficient backend that reaches close to one million environment steps per second for training classic RL agents. The exact same environments are directly playable by humans in a browser, making them also suitable for cognitive science studies.
By accelerating the development of novel environments
from conception to implementation ready for efficient RL training, 
PlayTrain allows researchers to shape environments around their research questions, rather than limiting those questions to existing environments---opening new avenues for RL research. \looseness=-1 

\subsection*{AI use statement}
In this work, we used generative AI tools for implementing methods.
We have not used generative AI tools for generating synthetic data sets,
developing theoretical models or conceptual frameworks, formulating mathematical claims, 
providing critical ingredients for proving mathematical claims, 
assisting in the writing of proofs, 
proposing or refining hypotheses,
designing or providing feedback on research methodology or experiments,
cleaning and reformatting datasets, 
supporting qualitative and thematic data analysis, 
or interpreting results, and assistance with translation is not applicable to this work.
Additionally, we used generative AI tools for creating and editing software code and Markdown prompt files, 
creating and modifying scientific figures, identifying related literature, 
and copy-editing the text for readability.
We have reviewed all AI-assisted work. LLM-generated code and Markdown prompt files were verified and tested for correctness by the first author.
We take responsibility for the final content of this work, including text, claims, or artifacts produced with the aid of generative AI.

\subsection*{Ethics statement}
The human-play study was approved by our institution's Institutional Review Board. 
Participants were recruited on Prolific, gave informed consent, and were compensated for their time; no personally identifiable information was
collected. 
All other experiments use simulated environments and involve no human subjects or sensitive data.

\subsection*{Reproducibility statement}
Implementation details are given in Appendices~\ref{app:bench}, \ref{app:envcost}, and~\ref{app:trainer}. The full generation prompt is in Appendix~\ref{app:prompt}, anonymized human data, source code and generated environments are provided in our repository, and figures/tables are reproducible with a bash command in the repository as well.

\ificlrfinal
\section*{Acknowledgments}
The authors are grateful for support from the Kempner Institute for the Study of Natural and Artificial Intelligence at Harvard University.
Kazuki Irie is grateful for support from the Wu Tsai Institute at Yale University.
\fi

\bibliography{iclr2027_conference}
\bibliographystyle{iclr2027_conference}

\appendix
\clearpage

\section{Further Discussions and Related Work}
\label{app:discussion}

\subsection{Related Work}
\label{app:related}
Traditional research in RL has typically held the environment as a fixed backdrop where agents are both trained and evaluated on, bypassing the core challenge of generalization. 
One approach to address this limitation is the use of procedural generation \citep{justesen2018illuminating, cobbe2020procgen}, allowing us to generate variations of environments along some pre-specified axes,
introducing a proper train/test split for evaluating RL agents.

More recently, advances in large generative models have shown  promising results in modeling entire environments using a neural network.
For example, the Genie model series \citep{bruce2024genie, genie2} are trained as a predictive world model on a large amount of game-playing videos with a learnable latent action space, so once they are trained, they can sequentially generate a pixel-level observation as a response to a discrete action, effectively simulating an environment.

Another line of work trains generative models of game code (essentially a neuro-symbolic approach).
For instance, OMNI-EPIC \citep{faldor2024omniepic} prompts an LLM to generate code that defines new PyBullet-based 3D environments with their accompanying reward functions. These code environments are conditioned on the agent's past performance in order to continually propose tasks at the frontier of what the agent can currently learn. DiCode \citep{mitsides2026dicode} applies the same idea in Craftax.
EnvGen \citep{zala2024envgen} follows a similar logic by using an LLM to generate environment \textit{configurations} for a preexisting simulator (e.g., Crafter). EnvGen then iteratively modifies those configurations to train an agent on specific tasks it struggles with.\looseness=-1

Both these lines of methods automating generation of environments---either directly on the pixel level or through engine-confined code---have fundamental limitations though. For example, a Genie environment is fully encoded in weights of a neural network, and therefore, cannot be inspected, edited, or replayed deterministically. 
OMNI-EPIC and EnvGen are confined by a physics engine and a configuration space. 
In contrast, PlayTrain is bounded by neither of these issues: every environment stays inspectable, editable, and deterministic, while it is not confined to specific engines.

Closely related to our approach, the AI GameStore \citep{ying2026aigamestore} generates games using LLMs based on popular app-store titles, and turns them into an open-ended benchmark for \textit{evaluating} LLM agents. 
However, AI GameStore is a benchmark rather than a training framework, and its games run far under practical RL throughput in terms of speed. PlayTrain overcomes this speed challenge and make that class of games fast enough to \textit{train} on.

In fact, recent development of RL environments has also focused on improving \textit{speed}. 
In particular, JAX \citep{jax2018github} and the subsequent wave of GPU-vectorized simulators \citep{brax, gymnax, jumanji} carried that processing of environments onto the GPU.
Beyond GPU vectorization, speed has been pursued in 2 levels: through \textit{engines} and \textit{systems}.
For engines, CuLE \citep{dalton2020cule} ported the Arcade Learning Environment to CUDA, removing CPU-to-GPU transfers and emulated thousands of Atari environments in parallel.
OCTAX \citep{radji2025octax} and PuzzleJAX \citep{earle2025puzzlejax} do the same for CHIP-8 and PuzzleScript, vectorizing those engines in JAX (note that using PlayTrain, engines built or emulated in JavaScript, such as PuzzleScript, CHIP-8, Pico-8, can become RL environments as they are). Madrona \citep{shacklett2023madrona} instead builds a \textit{custom GPU-native engine} expressive enough to host hand-written environments on the GPU. 

At the system level, Sample Factory and SEED RL decouple acting from learning to make full use of a node or actor fleet \citep{petrenko2020samplefactory, espeholt2019seed}, and EnvPool batches hand-written C++ environments on an in-process threadpool \citep{weng2022envpool}. 
PufferLib combines both, pairing asynchronous vectorization with its own trainer and runs pixel benchmarks such as Atari and Procgen. Their headline speeds comes from \textit{Ocean} though, their hand-written C environments with state vector observations  \citep{suarez2024pufferlib, suarez2025pufferlib2}.

These system and engine level approaches have trade-offs though. 
Game engines make mechanics and parameters difficult to modify, while system-level optimizations are mostly limited to existing environments that output symbolic states instead of images. 
PlayTrain avoids engines entirely by writing environments as standard JavaScript programs, using a fast system architecture to deliver high-speed pixel observations for reinforcement learning.

\subsection{Further discussions}
\label{app:furtherdisc}

\textbf{Scaling the number of environments: meta-RL, co-evolution, open-endedness.}
Certain RL methods may largely benefit from the potential of PlayTrain to generate and scale the number of diverse RL environments one can train an agent on.
In particular, the main bottleneck of certain meta-RL methods or learning-to-learn RL algorithms \citep{schmidhuber1987evolutionary, schmidhuber1998reinforcement}, such as \citet{oh2025discovering}'s, is the data (i.e., the scale and diversity of training environments) rather than the algorithm itself.\looseness=-1

PlayTrain may also be useful to advance co-evolution methods \citep{hillis1990parasites, sims1994competition, rosin1997competitive,blair1997makes}, such as POET \citep{wang2019poet}, and unsupervised environment design \citep{dennis2020paired, jiang2021plr, parkerholder2022accel}.
For example, a POET-styled environment commits itself to the variables it mutates before a simulator exists, and can only mutate what is exposed.
In PlayTrain a prompt edit can expose new mutation variables, or a brand-new environment outright. 
That edit happens once, outside the search loop. 
Mutating along those variables then costs the same as it does for POET.
More generally, PlayTrain may also serve as a tool to continually generate diverse environments and train agents for open-ended learning \citep{openendedteam2021xland}. PlayTrain itself may be part of the training/evolution loop in such a machine learning paradigm.

\textbf{Facilitating comparison with VLM game-playing agents.}
There has been an increasing interest in evaluating abilities of large visual language models (VLMs) as game-playing agents.
PlayTrain can also contribute to such research, as the exact same browser-based JS games can now be played by a VLM, a conventional RL agent, or a human player, making their scores directly comparable.

\textbf{Creating novel environments to fundamentally advance RL \& Evaluation challenges.}
PlayTrain largely facilitates development of novel environments for RL research.
This opens up many promising directions in fundamental RL research, which have traditionally been difficult due to lack of appropriate environments.
This may include development of diverse (partially obervable) environments with hard-exploration or specific-memory/cognition challenges, enabling research to advance generalization of exploration and memory algorithms, respectively.
One remaining challenge in automation is the evaluation of created novel environments, since the definitive evaluation of new environments would require test-playing by humans, at least as of today.

\begin{figure}[t]
\centering
\includegraphics[width=\linewidth]{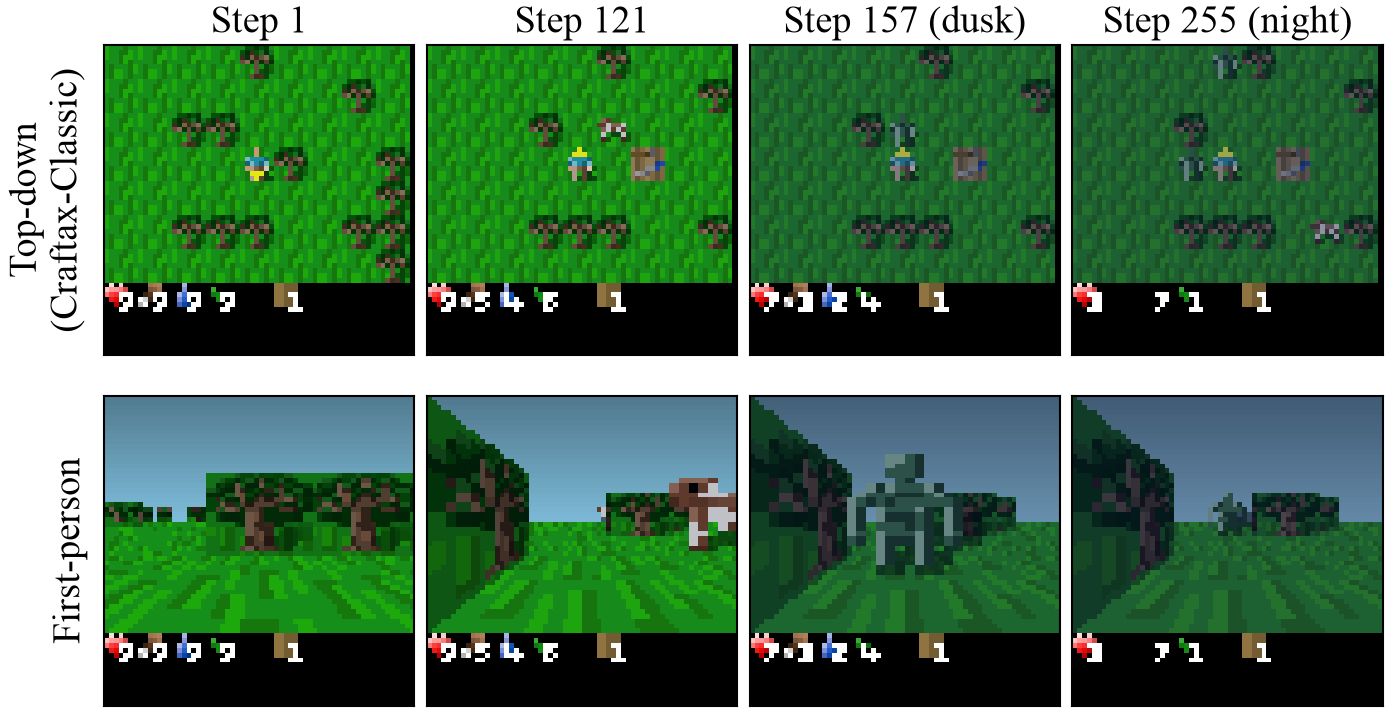}
\caption{Our JavaScript implementation of CraftAx-Classic and its first-person variation, only differing in observations. The underlying gamestate in each column are identical.}
\label{fig:craftax_views}
\end{figure}

\textbf{From single-file to multi-file.}
PlayTrain's pipeline is intentionally designed to generate and refine single-file JS environments for simplicity but is not restricted to this single-file setup. 
As environments grow richer and demand more complexity, generating multiple files makes their dynamics easier to manage and capture. 
For instance, 3D environments are far better modularized across multiple files for both human and machine legibility.

As a simple demonstration of this multi-file setting, we created an \textit{identical} JavaScript port of CraftAx-Classic's \texttt{C} implementation in PufferLib \citep{suarez2025pufferlib2} and built a first-person variation of it, both of which were designed to be compatible and playable in PlayTrain (Figure~\ref{fig:craftax_views}). 
The process in which this port and its variant were generated involved looping a Claude Code agent to build a port and running subsequent tests that step the JS port alongside the \texttt{C} implementation to check that the game states were the same. 
The generated variant was built on top of the JS implementation by only changing observations by adding voxel rendering to PlayTrain's existing rasterizer for the first-person view. Both environments are available and playable in our repository.

\section{PlayTrain Action Space and Observations}
\label{app:pt_act_obs}

We describe here the default eight-action space, how the backend turns actions into keyboard states a game can receive, how a different action space is defined in a JSON file including continuous ones, and what a single step returns in terms of observations.
PlayTrain is intentionally designed so that the action space is defined by the framework's backend instead of the games that are generated by it.
In doing so, the generated games can share the same action space and allow agents to train across games.
This section goes into detail about how it all works. 


\begin{table}[H]
\centering
\caption{The default \texttt{Discrete(8)} action space.}
\label{tab:action-space}
{\footnotesize
\begin{tabular}{@{}llll@{}}
\toprule
Index & Name & Keycodes & Delivery \\
\midrule
0 & NOOP    & ---    & --- \\
1 & LEFT    & 37     & held \\
2 & RIGHT   & 39     & held \\
3 & UP      & 38     & held \\
4 & DOWN    & 40     & held \\
5 & D       & 32     & press \\
6 & LEFT+D  & 37, 32 & held + press \\
7 & RIGHT+D & 39, 32 & held + press \\
\bottomrule
\end{tabular}}
\end{table}

\textbf{Action Space. } PlayTrain's default action space is \texttt{Discrete(8)} and is utilized across all the games presented here. 
In PlayTrain, the agent selects an action among the 8 in \texttt{Discrete(8)} by emitting an integer. 
From that emitted integer, the PlayTrain backend transforms it into the keycodes
listed in Table~\ref{tab:action-space}, handling whether each key is held, pressed, or both.
Once the keyboard state is handled by the PlayTrain backend, the game's \texttt{draw()} function is called. 
The game then updates its objects and issues its drawing commands, and our rasterizer writes the
resulting frame, which becomes the next observation.
Under \texttt{NOOP} the backend holds no keys, yet \texttt{draw()} still runs and the agent still
receives a new observation.

\textbf{Observation. } A game's canvas size is independent of the observation since our rasterizer
runs drawing commands at the observation's resolution regardless.
By default, observations are set a single 64$\times$64 RGB frame with no frame stacking.
Table~\ref{tab:step-return} lists everything a step returns.

\begin{table}[H]
\centering
\caption{What a step returns. \texttt{terminated} says the episode ended; \texttt{info.gameState} says how.}
\label{tab:step-return}
{\footnotesize
\begin{tabular}{@{}lll@{}}
\toprule
Field & Type & Meaning \\
\midrule
observation           & \texttt{uint8[64,64,3]} & the frame our rasterizer produced this step \\
reward                & float                   & change in the game's score \\
terminated            & bool                    & true when \texttt{gameState} is no longer \texttt{PLAYING} \\
truncated             & bool                    & true at \texttt{max\_steps}, default 2000 \\
\texttt{info.score}     & float                 & cumulative score \\
\texttt{info.lives}     & float                 & lives remaining. Reaching 0 triggers \texttt{GAMEOVER} \\
\texttt{info.gameState} & string                & \texttt{PLAYING}, \texttt{WIN}, \texttt{GAMEOVER}, or \texttt{EXIT}\\
\texttt{info.seed}      & int                   & seed for the current episode \\
\bottomrule
\end{tabular}}
\end{table}

\begin{figure}[H]
\centering
\begin{minipage}{0.62\linewidth}
\begin{lstlisting}[style=jsonfig]
"default8": [
  { "name": "NOOP",   "held": [],   "press": null },
  { "name": "LEFT",   "held": [37], "press": null },
  { "name": "LEFT_D", "held": [37], "press": 32   },
  /* [...] five more */
],
"mouse2d": { "type": "box",
  "channels": ["pointer_x", "pointer_y", "button:mouse"] }
\end{lstlisting}
\end{minipage}
\caption{Two entries from the action-space file, \texttt{default8} abridged. \texttt{mouse2d} gives a continuous action space.}
\label{fig:action-spaces}
\end{figure}

The action space we use is a named entry in a JSON configuration file and is not built into PlayTrain.
Each named entry is simply a list of actions, and each of those actions names the keys to hold down and an optional additional key to press.
A held key is down for the step, while a press key is down for the step \textit{and} fires \texttt{keyPressed()} once.
An entry can instead declare channels, enabling an agent to use a continuous space.
This flexiblity means that anyone else can setup their own specified action-space by creating an entry of their own. 

A continuous action-space (for settings such as a mouse or gamepad) works the same way, except that the backend sets \texttt{mouseX}, \texttt{mouseY}, and \texttt{mousePressed} instead of key state. 
For example, continuous input values are given to the PlayTrain backend as integers, while the PlayTrain backend transforms those integers as \texttt{mouseX}, \texttt{mouseY} and \texttt{gamepadAxes} values the games can read. Integers are used because floating point values are not exactly reproducible across JavaScript engines. 

\textbf{Configuration. } PlayTrain's default configurations are designed to be easily adaptable.
On the observation side users can adjust the resolution, set the frame skip, render skip, and frame stack, change the truncation horizon, or modify the RGB settings (e.g. grayscale).
On the environment side users can set the number of environments and worker threads, choose/make their own the action space, and select how episodes are seeded. 
For instance, a training run can set a fresh seeds for each episode, a fixed seed across episodes, or only select seeds drawn from a defined pool.
None of these changes require changing a game file.

\section{Implementation Details}
\label{app:impl}
This section covers three things: how exactly a PlayTrain environment runs, how the same games run on the browser and the Node backends baselines we compare against, and how exactly IMPALA and PPO trainers are built.

\subsection{Environment Implementation}
\label{app:backend}

\begin{table}[h]
\centering
\small
\setlength{\tabcolsep}{0.5em}
\begin{tabular}{@{}l>{\raggedright\arraybackslash}p{0.76\linewidth}@{}}
\toprule
Group & Commands \\
\midrule
Canvas \& frame & \texttt{createCanvas}, \texttt{background} \\
Color state & \texttt{fill}, \texttt{stroke}, \texttt{noFill}, \texttt{noStroke}, \texttt{strokeWeight}, \texttt{color}, \texttt{lerpColor} \\
Primitives & \texttt{rect}, \texttt{ellipse}, \texttt{circle}, \texttt{arc}, \texttt{triangle}, \texttt{quad}, \texttt{line} \\
Modes & \texttt{rectMode}, \texttt{ellipseMode} \\
Transforms & \texttt{push}, \texttt{pop}, \texttt{translate}, \texttt{rotate}, \texttt{scale} \\
Custom shapes & \texttt{beginShape}, \texttt{vertex}, \texttt{endShape} \\
Offscreen & \texttt{createGraphics}, \texttt{image} \\
Input & \texttt{keyIsDown}, \texttt{keyPressed}; \texttt{mouseX}, \texttt{mouseY}, \texttt{mouseIsPressed}; \texttt{gamepadAxes} \\
Text & \texttt{textSize}, \texttt{textAlign}, \texttt{text} \\
\bottomrule
\end{tabular}
\caption{37 p5.js commands the C\texttt{++} later binds to. The command names match p5.js so that a generated file runs as written. Text commands never appears in observations, and operations such as noLoop, frameRate, cursor, and more a compatible no-ops so that games don't crash.}
\label{tab:p5-subset}
\end{table}

Here we detail our QuickJS implementation its wiring to PlayTrain: how it gets compiled, how a game's drawing commands get to our custom rasterizer, and what one step actual performs.
The key element to PlayTrain's speed comes from QuickJS: it enables the same-program design, where an environment runs inside the trainer's own process, and most of the other backend decisions follow from it.

\begin{table}[h]
\centering
\small
\begin{tabular}{@{}lll@{}}
\toprule
Piece & Replaces & Why \\
\midrule
QuickJS (C) & V8, in Node or a browser & embeds in-process\\
p5 layer (C\texttt{++}) & the \texttt{p5.js} library & draw calls land in compiled code \\
Rasterizer (Rust) & the browser canvas & writes straight into the observation buffer \\
Frozen math (C) & the platform's \texttt{libm} & identical \texttt{sin} and \texttt{cos} everywhere \\
\bottomrule
\end{tabular}
\caption{The four pieces compiled into the backend, in the order a frame passes through them.}
\label{tab:backend-pieces}
\end{table}

\textbf{Compilation.} The PlayTrain backend itself compiles once per machine it runs on. Running a PlayTrain environment loads that build and reads the game's JavaScript, which the engine interprets unless that game has been compiled ahead of time as described below.
The compilation process involves four pieces: QuickJS, p5 C++ layer, Rust rasterizer, and frozen math.
As a primer, JavaScript by itself is just text on a screen.
In order to turn the text into a program, a JavaScript \textit{engine} is required. 
This engine is commonly V8 and often embedded in Node.js.
That V8 engine is heavily optimized for speed with advanced JIT compilation, and previous work involving JS RL runtimes uses it \citep{openai2016universe, shi2017wob}.
But V8's machinery is heavy, complex, and inflexible.
QuickJS (QJS), an alternative JS engine, is what we use instead.
Interestingly enough, it is not a faster engine --- just one built as a C library, and actually is many times slower than the V8 engine packaged in Node.
QJS only works for us because it (1) easily embeds directly into PlayTrain, (2) is very cheap to initialize, and (3) doesn't require any additional runtime machinery.

\textbf{Ahead-of-time compilation.} 
A game's JavaScript is compiled separately from the backend ahead of time into an executable form that a C compiler treats as read-only data. 
This lets our JS games load directly instead of parsing the JS source code. 
This is performed by default for every game and every training run presented in this paper uses the compiled build. 
If a game has \textit{not} been compiled yet, such as a freshly generated game/variant, the game runs through the QuickJS interpreter until the game's compiled build finishes. 

The backend itself is built with profile-guided and link-time optimization for the machine it runs on.
At run time we also skip rasterizing any frame whose drawing commands are unchanged.
The game file is never modified and the frames are identical either way, so only the speed of the game's runs change.

\textbf{QuickJS.} Our QJS is a fork of Bellard's QuickJS extended with tail-call dispatch and ahead-of-time compilation adjustments, and the engine itself contains six C files: \texttt{quickjs.c}, \texttt{dtoa.c}, \texttt{libregexp.c}, \texttt{libunicode.c}, \texttt{cutils.c} and \texttt{quickjs-libc.c}.
(Tail-call dispatching just means that each instruction jumps straight to the next rather than returning to a central dispatcher).
The engine is driven by a simple C interface, and this interface travels in both directions: PlayTrain can also register its own C\texttt{++} functions as ordinary JS globals, meaning that when a game calls the \texttt{ellipse} drawing command, the interpreter jumps \textit{directly} into PlayTrain's pre-compiled rendering code rather than a JS library.
That pre-compiled rendering code \textit{is} our custom C\texttt{++} p5 layer mentioned in the main text.
That layer communicates with our Rust rasterizer through a C ABI, which is simply the conventional C calling format that both a compiled C\texttt{++} and a compiled Rust program can read.

\textbf{Rasterization.} The rasterizer in itself has its own trajectory when being built for PlayTrain.
At the beginning of PlayTrain's development, we used node-canvas, and then built our own rasterizer in plain JavaScript (\texttt{raster.mjs}).
This version matched node-canvas's interface and buffer format except for the omission of Anti-Aliasing (AA)---building a rasterizer that utilized AA added more friction since different browser versions and OSes define AA differently.
That specific \texttt{raster.mjs} file is still used as the rasterizer deployed for actual browser playtesting, but we later used it as the reference for our eventual Rust rasterizer port, which now all training uses.
And because it is interlocked with the p5 C\texttt{++} API layer, the low-level rasterizer enabled even greater speedups.
For the games presented in this paper, that low-level rasterizer draws the game's canvas into the observation buffer at the default $64{\times}64$ resolution.
Consequently, there is no full-size render nor downscale pass,
The rasterizer also defines its own \texttt{sin} and \texttt{cos} since Rust's differ slightly from the JS V8 engine.
We build test scripts to make sure the \texttt{raster.mjs} and the native Rust rasterizer produce identical observations across the entire game suite.

Lastly, we freeze math functions since some games use them for game logic operations.
We therefore fork \texttt{OpenLibm} (a portable open-source version of the math library \texttt{fdlibm}) for their math functions, such as \texttt{sin} and \texttt{cos}.

Taking everything together, one step reduces to four operations:
\begin{enumerate}
\item \texttt{draw()} runs once.
\item Its drawing commands land in the precompiled C\texttt{++} p5 library.
\item The rasterizer writes the resulting pixels straight into the observation buffer.
\item The reward is the score delta; termination is read from the game state.
\end{enumerate}

\subsection{Other Backend Implementations}
\label{app:playwright_node_qjs}

In this section we detail the exact methods for Figure~\ref{fig:env_efficiency}B for the Playwright and the Node/V8 categories.
In this figure, 3 different backends were used: a headless browser through the Playwright API, a browserless node instance running on the V8 engine, and the QuickJS that the paper relies on.
The game's JavaScript is identical in all three; only the layers beneath it change.

For our Playwright backend comparison, we don't use the native custom rasterizer or p5 C\texttt{++} layer, as those two parts are not compatible with a browser. 
Instead, all of the browser mumbo-jumbo is present and driven headlessly through the Playwright API. For our purposes we use the Chromium build that ships with \texttt{playwright-core} 1.57.0. 
The p5 layer is now just the actual p5.js library package. 
As for the stepping process, each \texttt{step()} runs a \texttt{page.evaluate} call from the Playwright API. 
A \texttt{draw()} is then called, leading the resulting frame to get copied from the \texttt{getImageData} command (an HTML Canvas 2D API). 
This copied frame gets downsampled to the trainer's observation resolution and then moves back into the Node  through base64 encoding.

On the headless Node V8 engine backend, the browser is no longer participating. This usually isn't
possible, as browser JS code (like \texttt{p5.js}) typically needs browser machinery such as the
DOM and canvas in order to run.
Our strategy (similar to what is presented in the main text) forces the JS browser code to instead speak to our own API layer that directly connects to our p5.js JS rewrite.

This rewritten JS version of p5.js then runs drawing commands using our rasterizer (that is compiled to WebAssembly) in order to generate pixels \textit{outside} the browser. 
Each generated pixel turns to frames that then move from Node into Python through a pipe and shared memory. Python drives the stepping.

\begin{table}[h]
\centering
\caption{The three backends behind Figure~\ref{fig:env_efficiency}B. The game's JavaScript code is identical across all three. The browser backend cannot use our p5 layer or rasterizer so it runs real p5.js on the browser canvas and returns observations through the Playwright API.
}
\label{tab:backend-ladder}
\small
\begin{tabular}{@{}llll@{}}
\toprule
 & Browser (Playwright) & Node/V8 & QuickJS (PlayTrain) \\
\midrule
JS engine       & V8 in Chromium   & V8                & QuickJS, linked in-process \\
p5 layer        & real p5.js       & our rewrite (JS)  & our rewrite (C\texttt{++}) \\
Rendering       & browser canvas   & rasterizer (wasm) & rasterizer (native) \\
Observation out & base64           & pipe to Python    & into the obs buffer \\
Driver          & Node             & Python            & C \\
\bottomrule
\end{tabular}
\end{table}

\subsection{Implementation Details of Trainers}
\label{app:trainer}

\begin{figure}[H]
\centering
\begin{tikzpicture}[x=1cm,y=1cm,
  blk/.style={draw=black!60, line width=0.4pt, minimum height=0.42cm, anchor=west, font=\scriptsize, inner sep=0.5pt},
  env/.style={blk, fill=orange!35},
  gpu/.style={blk, fill=blue!25},
  cpuf/.style={blk, fill=blue!12},
  chan/.style={blk, fill=black!12},
  waitb/.style={blk, pattern=north east lines, pattern color=black!45},
  lab/.style={font=\scriptsize},
  lane/.style={anchor=east,font=\footnotesize}]

\node[anchor=west,font=\footnotesize\bfseries] at (0,1.6) {(A) Shared-CPU actors};
\node[lane] at (-0.15,0.8) {actor (CPU)};
\node[lane] at (-0.15,0.0) {GPU};
\foreach \i in {0,1,2,3,4} {
  \pgfmathsetmacro{\xa}{\i*1.7}
  \pgfmathsetmacro{\xb}{\xa+0.7}
  \node[env,  minimum width=0.7cm] at (\xa,0.8) {step};
  \node[cpuf, minimum width=1.0cm] at (\xb,0.8) {forward};
}
\node[gpu, minimum width=8.5cm] at (0.0,0.0) {learner only, no inference};
\node[lab] at (4.25,0.0) {learner only, no inference};
\draw[->,black!70] (0,-0.55) -- (9.4,-0.55) node[right,font=\scriptsize]{time};

\begin{scope}[yshift=-3.0cm]
\node[anchor=west,font=\footnotesize\bfseries] at (0,1.6) {(B) Centralized batched inference};
\node[lane] at (-0.15,0.8) {actor};
\node[lane] at (-0.15,0.0) {server (GPU)};
\node[env,  minimum width=0.7cm]  at (0.0,0.8) {step};
\node[chan, minimum width=0.55cm] at (0.7,0.8) {send};
\node[waitb, minimum width=2.5cm] at (1.25,0.8) {};
\node[lab] at (2.5,0.8) {waiting};
\node[chan, minimum width=0.55cm] at (3.75,0.8) {reply};
\node[chan, minimum width=1.0cm]  at (1.25,0.0) {batch};
\node[gpu,  minimum width=1.5cm]  at (2.25,0.0) {forward};
\node[env,  minimum width=0.7cm]  at (4.3,0.8) {step};
\node[chan, minimum width=0.55cm] at (5.0,0.8) {send};
\node[waitb, minimum width=2.5cm] at (5.55,0.8) {};
\node[lab] at (6.8,0.8) {waiting};
\node[chan, minimum width=0.55cm] at (8.05,0.8) {reply};
\node[chan, minimum width=1.0cm]  at (5.55,0.0) {batch};
\node[gpu,  minimum width=1.5cm]  at (6.55,0.0) {forward};
\node[waitb, minimum width=1.25cm] at (0.0,0.0) {};
\node[lab] at (0.6,0.0) {idle};
\node[waitb, minimum width=1.75cm] at (3.75,0.0) {};
\node[lab] at (4.6,0.0) {idle};
\node[waitb, minimum width=1.2cm] at (8.05,0.0) {};
\node[lab] at (8.65,0.0) {idle};
\draw[->,black!70] (0,-0.55) -- (9.4,-0.55) node[right,font=\scriptsize]{time};
\end{scope}

\begin{scope}[yshift=-6.0cm]
\node[anchor=west,font=\footnotesize\bfseries] at (0,1.6) {(C) Vectorized worker, double-buffered};
\node[lane] at (-0.15,0.8) {env threads (C++)};
\node[lane] at (-0.15,0.0) {worker GPU};
\foreach \i in {0,1,2,3,4} {
  \pgfmathsetmacro{\xa}{\i*1.86}
  \pgfmathparse{mod(\i,2)==0 ? "orange!25" : "orange!60"}
  \edef\ca{\pgfmathresult}
  \pgfmathparse{mod(\i,2)==0 ? "step group 0" : "step group 1"}
  \edef\la{\pgfmathresult}
  \node[blk, fill=\ca, minimum width=1.8cm] at (\xa,0.8) {\la};
}
\foreach \i in {0,1,2,3,4} {
  \pgfmathsetmacro{\xa}{\i*1.86}
  \pgfmathparse{mod(\i,2)==0 ? "blue!45" : "blue!18"}
  \edef\cb{\pgfmathresult}
  \pgfmathparse{mod(\i,2)==0 ? "infer group 1" : "infer group 0"}
  \edef\lb{\pgfmathresult}
  \node[blk, fill=\cb, minimum width=1.8cm] at (\xa,0.0) {\lb};
}
\draw[->,black!70] (0,-0.55) -- (9.4,-0.55) node[right,font=\scriptsize]{time};
\end{scope}
\end{tikzpicture}
\caption{Three inference configs. \textbf{(A)} Policy on the CPU per actor: stepping and inference
use the same cores. \textbf{(B)}  Central GPU thread: actors sit idle waiting for their action, the server between batches. 
\textbf{(C)} Double-buffering: two groups of environments alternate. One group steps while the other's inference runs.}
\label{fig:trainer_timeline}
\end{figure}
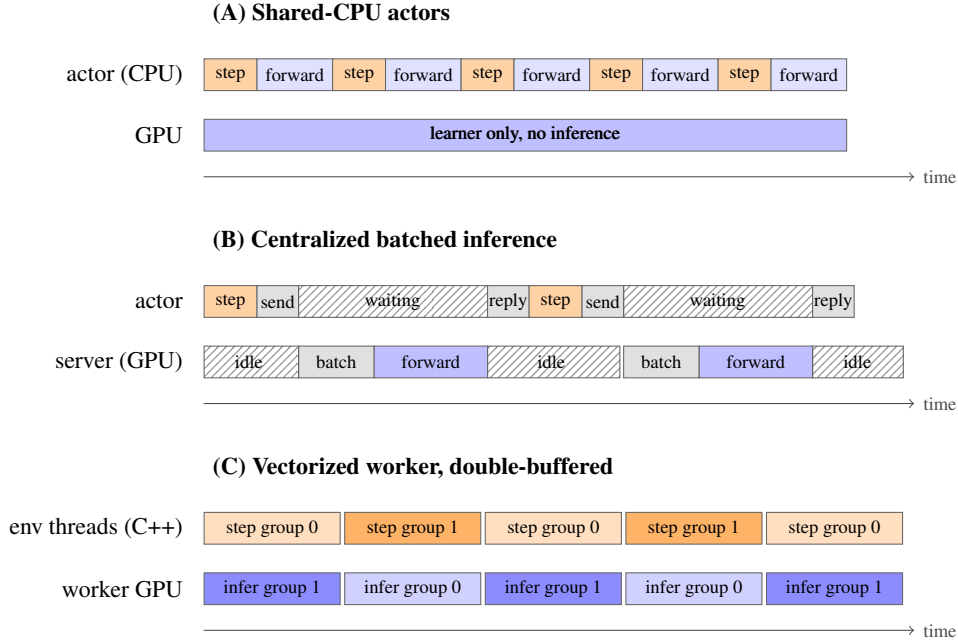

Much of our efforts in building PlayTrain required porting IMPALA and PPO into our trainer system. 
IMPALA specifically was the key into enabling our throughput benchmarks and porting it into our PlayTrain ecosystem involved referencing many sources. 

We detail that process here and are enthusiastic about pushing more IMPALA improvements in the future. Both trainers PPO and IMPALA utilize the same C++ vector environment. 

\textbf{IMPALA}. Our re-implementation of IMPALA \citet{espeholt2018impala} heavily referenced the polybeast and torchbeast architecture \citep{kuttler2019torchbeast} but pulls from Sample Factory \citep{petrenko2020samplefactory} for added modern improvements.
More specifically, we derive details such as V-trace, losses, and buffer layouts from torchbeast and build tests to make sure that our version's outputs are identical. 
Added improvements to the \textit{trainer} came from polybeast/SEED RL's centralized batched inference \citep{espeholt2019seed} and Sample Factory's double-buffered environment sampling. 
Our IMPALA trainer is written in Python and PyTorch without any other language dependencies.  

We initial adopted monobeast's \texttt{shared\_cpu} design in the process of porting IMPALA for PlayTrain. 
Within a \texttt{shared\_cpu} design, each actor has its own environment and samples actions by running the policy network on the CPU per observation. 
The weights from the policy network reside in shared CPU memory, which every actor reads from while the learner writes its updates directly into it. 
While this setup is memory-safe and simple, our environments were simply too fast in comparison to the trainer's top speeds. 

To address the speed asymmetry, we looked at SEED RL and polybeast \citep{espeholt2019seed, kuttler2019torchbeast}. 
In their work, the key contribution was centralized inference and involved actors no longer running the policy network/model. 
Actors instead only receives actions, writes observations, and waits for the next action. 
A \textit{centralized inference thread} now takes care of the forward pass actors used to handle in the previous \texttt{shared\_cpu} design: 
the thread collects and batches observations from actors waiting for actions and runs a forward pass on those batched observations.
The thread finally sends actions back to to the waiting actors afterward. 
Another speed issues arises, however.
Because the forward pass is disjointed from the actor, there is latency between actors stepping their environments and receiving the next actions. 
There are explicitly four stages of latency: \texttt{send}, \texttt{batch}, \texttt{forward}, and \texttt{reply}. 

To remove this latency, we have to give back the forward pass to the actor in a diferent way.
We can now instead transform actors into \textit{workers} and let each own its \textit{own} GPU copy of the policy network and 256 environments in one C++ vector environment \citep{weng2022envpool}.
This setup enables each worker to infer its own batch environments and step \textit{all} of their environments in a single call. 

Only two costs remain after this, which are now the stepping and inferring. Usually, these two processes cannot occur at the same time because both rely on one another within a worker.
Yet, by using the double buffering method \citep{petrenko2020samplefactory}, we can let two groups of environments run at the same time as mentioned in the main text (Figure~\ref{fig:trainer_timeline}B, Table~\ref{tab:dbuf-ablation}).

We now lastly mention the learner element in IMPALA. 
Our IMPALA learner continuously updates on batches of trajectories collected by the workers, using the V-trace loss. 
In our setup specifically, four GPUs in the node are split between learner and inference and it changes depending on the encoder being used (Table~\ref{tab:hyperparams}). 
If the IMPALA-CNN is used, two GPUs trainer the network with DDP while the other two hold the worker's copies of the policy network. 
If the Nature-CNN is used, one GPU trains as the learner while the remaining three are used for inference.

\begin{table}[t]
\caption{Training configuration. Each config is identical for every game. The two trainers are not matched on every property: IMPALA clips rewards to $\pm1$ and discounts at $0.99$ while PPO doesn't clip rewards and discounts at $0.999$. Lastly, the throughput-side elements differ as well, shown in the table. Topology and environment-count rows only apply to IMPALA.}
\label{tab:hyperparams}
\centering
\small
\begin{tabular}{lr@{\hskip 2.5em}lr}
\toprule
IMPALA / V-trace & & PPO & \\
\midrule
encoder & IMPALA-CNN & encoder & IMPALA-CNN \\
feature dim & 256 & environments & 192 \\
recurrence & none & rollout length & 128 \\
observation & $3\times64\times64$ RGB & minibatches & 8 \\
frame skip / stack & 1 / 1 & epochs per batch & 3 \\
batch size ($=M$) & 256 & learning rate & $2.5\times10^{-4}$, annealed \\
unroll length & 64 & discount & 0.999 \\
discount & 0.99 & GAE $\lambda$ & 0.95 \\
baseline cost & 0.5 & clip coefficient & 0.2 \\
entropy cost & 0.01 & value coefficient & 0.5 \\
reward transform & clip to $\pm1$ & entropy coefficient & 0.01 \\
gradient-norm clip & 40.0 & gradient-norm clip & 0.5 \\
optimizer & RMSProp & optimizer & Adam \\
learning rate & $5\times10^{-4}$ & precision & fp32 \\
$\alpha$ / momentum / $\epsilon$ & 0.99 / 0 / $10^{-5}$ & \texttt{torch.compile} & off \\
precision & bf16, channels-last & & \\
\texttt{torch.compile} &  max-autotune-no-cudagraphs & & \\
topology (Nature-CNN) & 15 workers $\times$ 5 threads & & \\
 & 1 DDP + 3 inference GPUs & & \\
environments & $15\times2\times256=7{,}680$ & & \\
topology (IMPALA-CNN) & 12 workers $\times$ 5 threads & & \\
 & 2 DDP + 2 inference GPUs & & \\
environments & $12\times2\times256=6{,}144$ & & \\
per-game tuning & none & per-game tuning & none \\
\bottomrule
\end{tabular}
\end{table}

\textbf{PPO}. Our PPO implementation is not as complex as our IMPALA optimizations and simply references CleanRL \citep{huang2022cleanrl}.
Everything in PPO runs solely on a single process with one GPU all in a synchronous loop. 
None of IMPALA's optimizations are needed since a synchronous on-policy loop never divorces acting from learning. 
The PPO trainer simply loops through iterations. In each iteration, our PPO collects a rollout of 128 steps from all 192 environments. It then optmizes on those 24,576 timesteps for 3 epochs over 8 minibatches. 

It is important to note that PPO goes through 3 gradient computations per frame while IMPALA only goes through one. 
This is a \textit{plausible} explanation for why PPO is more sample efficient and slower compared to IMPALA for certain environments. 

\begin{table}[t]
\caption{Double buffering, measured on the same trainer and games with everything else fixed.
Agent-steps/s under IMPALA with the Nature-CNN at the encoder's topology in Table~\ref{tab:hyperparams}, fifteen workers.
Splitting the environments into two groups so that one steps while the other's inference runs
(Figure~\ref{fig:trainer_timeline}C) helps most on the slowest environments.}
\label{tab:dbuf-ablation}
\centering
\small
\begin{tabular}{@{}lrrr@{\hskip 1.6em}lrrr@{}}
\toprule
Game & Double & Single & Ratio & Game & Double & Single & Ratio \\
\midrule
miner & 969k & 465k & 2.08$\times$ & heist & 878k & 747k & 1.18$\times$ \\
leaper & 937k & 488k & 1.92$\times$ & frostbite & 888k & 770k & 1.15$\times$ \\
coinrun & 960k & 501k & 1.92$\times$ & breakout & 932k & 809k & 1.15$\times$ \\
qbert & 860k & 465k & 1.85$\times$ & freeway & 1.04M & 904k & 1.15$\times$ \\
fruitbot & 796k & 436k & 1.83$\times$ & asteroids & 904k & 800k & 1.13$\times$ \\
chaser & 983k & 544k & 1.81$\times$ & starpilot & 904k & 816k & 1.11$\times$ \\
jumper & 904k & 511k & 1.77$\times$ & ninja & 917k & 868k & 1.06$\times$ \\
climber & 655k & 416k & 1.58$\times$ & bigfish & 904k & 878k & 1.03$\times$ \\
dodgeball & 898k & 586k & 1.53$\times$ & bossfight & 904k & 885k & 1.02$\times$ \\
maze & 973k & 659k & 1.48$\times$ & seaquest & 894k & 878k & 1.02$\times$ \\
caveflyer & 878k & 596k & 1.47$\times$ & plunder & 885k & 908k & 0.97$\times$ \\
space\_invaders & 1.01M & 829k & 1.22$\times$ & pong & 901k & 980k & 0.92$\times$ \\
\midrule
geometric mean & 904k & 673k & \textbf{1.34}$\times$ & \multicolumn{4}{r@{}}{median 1.20$\times$, range 0.92--2.08$\times$} \\
\bottomrule
\end{tabular}
\end{table}

\section{Full-Suite Learning Curves}
\label{app:suite}

\begin{figure}[H]
\centering
\includegraphics[width=\linewidth]{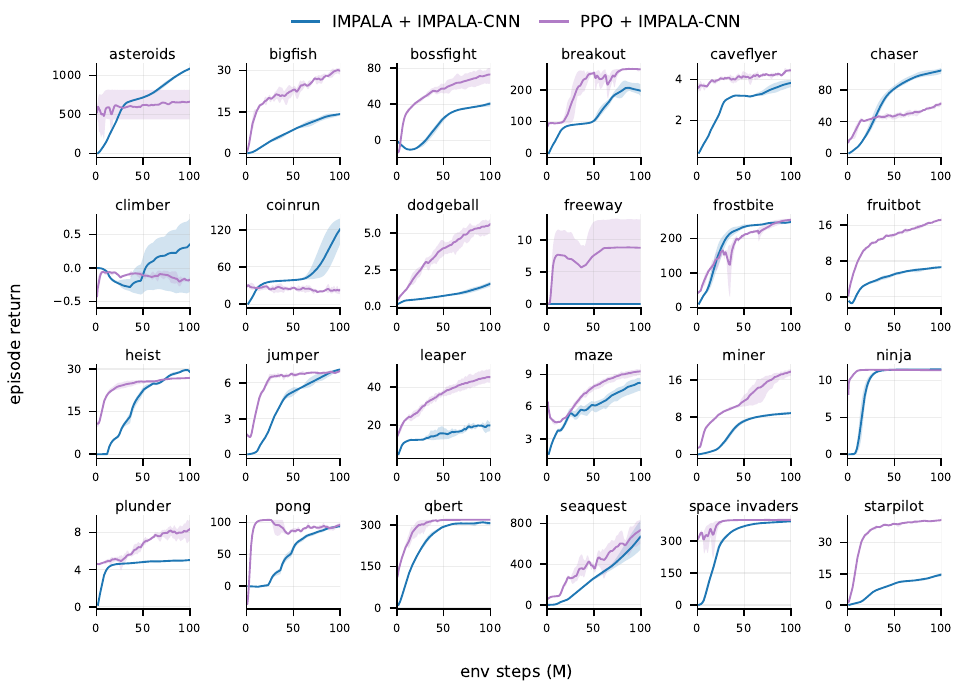}
\caption{IMPALA against PPO on all 24 games, both with the IMPALA-CNN encoder, 100M environment steps, three seeds per plot.
Lines are the seed mean, bands are the min and max. PPO curves start at 0.5M steps because only winning episodes terminated before then. \texttt{maze}, \texttt{heist}, and \texttt{freeway} have no failure state, so no episode ends until the 2{,}000-frame horizon at 12.3M steps, and before that point their IMPALA curves average over wins alone. 
}
\label{fig:suite_trainers}
\end{figure}

\begin{figure}[p]
\centering
\includegraphics[width=\linewidth]{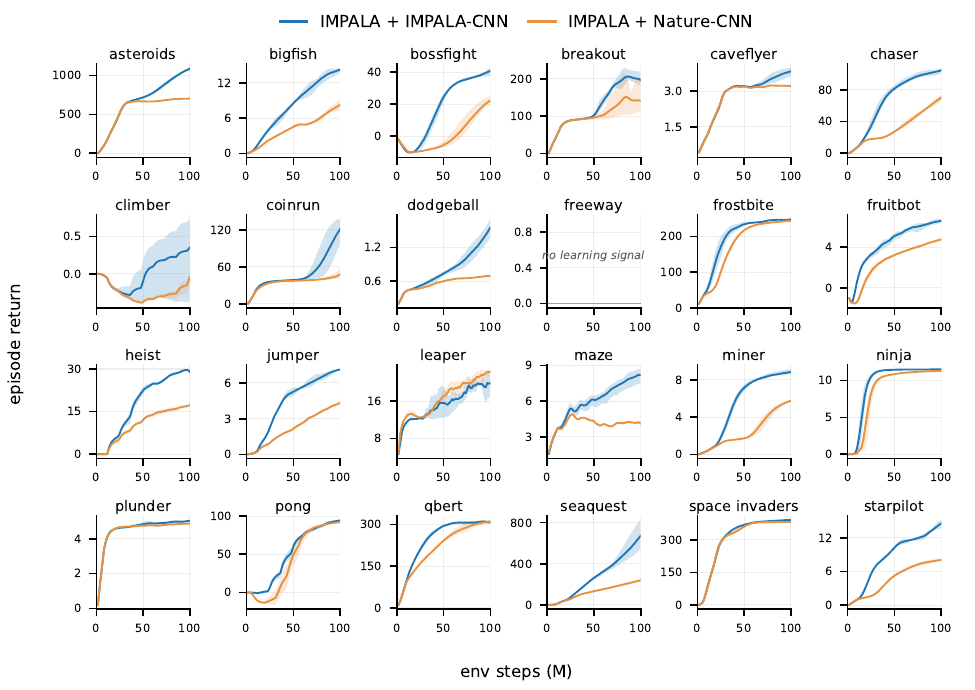}
\caption{IMPALA at both IMPALA-CNN and Nature-CNN encoders. Same setup as Figure~\ref{fig:suite_trainers}}
\label{fig:suite_enc_impala}
\end{figure}

\begin{figure}[p]
\centering
\includegraphics[width=\linewidth]{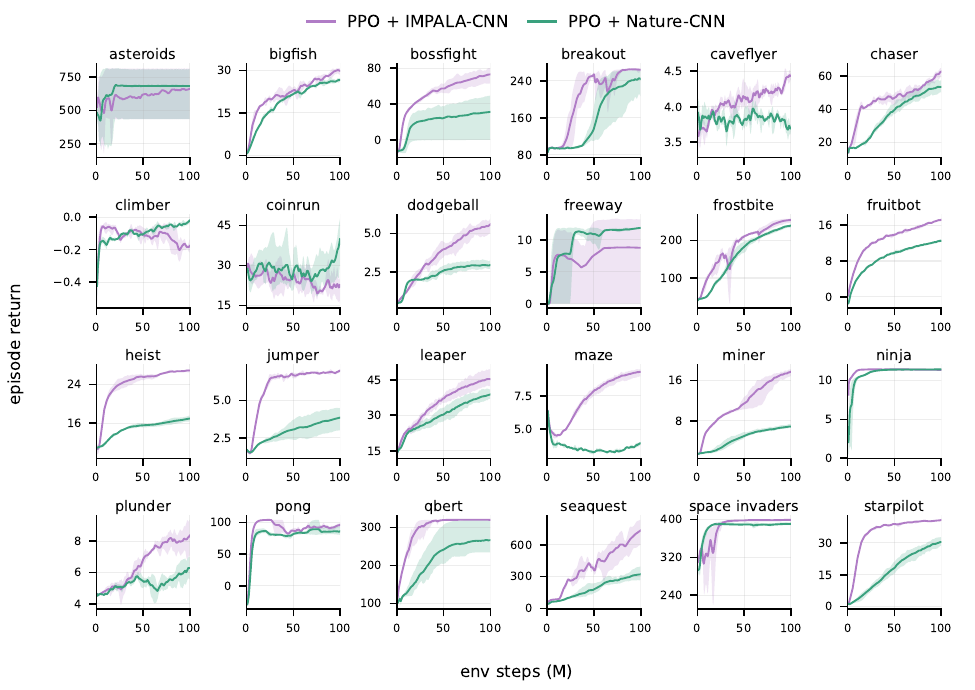}
\caption{PPO at both IMPALA-CNN and Nature-CNN encoders. Same setup as Figure~\ref{fig:suite_trainers}}
\label{fig:suite_enc_ppo}
\end{figure}

We present the learning curves of the 16 ProcGen and 8 ALE games across the IMPALA and PPO trainers.
We also test two encoders for each trainer: the IMPALA-CNN and Nature-CNN.
Each run lasted for 100M env steps.
Lines are the mean over 3 seeds and bands are their min and max.
PPO curves start at 0.5M steps because only wins have finished before the first truncation wave.
\texttt{maze}, \texttt{heist} and \texttt{freeway} have no failure state so their IMPALA curves before 12.3M steps average over finished episodes only.

IMPALA seems to fail to receive any reward at \texttt{freeway} but outperforms PPO on surprising titles such as \texttt{climber}, \texttt{coinrun}, \texttt{chaser}, \texttt{heist} and \texttt{asteroids}.
Its \texttt{freeway} zero holds across both encoders and all 3 seeds and matches the zero \citet{espeholt2018impala} report for IMPALA on ALE Freeway, while PPO reaches returns of 12 to 13 on five of six seeds.
On \texttt{climber} only IMPALA finishes above zero.
PPO is either equal or outperforms the remaining game titles and wins 17 of the 24 despite being the slowest arm to train (Table~\ref{tab:train-throughput}).
Table~\ref{tab:eval} reports final returns against random for every game.

\begin{table}[t]
\caption{Mean return over 8 held-out level seeds, averaged over three training seeds, for IMPALA and PPO with the IMPALA-CNN encoder after 100M steps (the runs in Figure~\ref{fig:suite_trainers}). R = random policy on the same seeds; the greedy final checkpoint is evaluated.}
\label{tab:eval}
\centering
\small
\begin{tabular}{lrrr@{\hskip 2.5em}lrrr}
\toprule
game & R & IMPALA & PPO & game & R & IMPALA & PPO \\
\midrule
asteroids & 513.8 & 1158.3 & 479.2 & heist & 2.5 & 10.2 & 4.0 \\
bigfish & 0.6 & 18.5 & 31.4 & jumper & 0.8 & 0.3 & 0.8 \\
bossfight & -13.2 & 71.2 & 76.3 & leaper & 12.5 & 12.0 & 23.6 \\
breakout & 80.0 & 263.3 & 254.6 & maze & 3.8 & 1.7 & 2.9 \\
caveflyer & 1.6 & 2.8 & 1.1 & miner & 2.1 & 8.0 & 11.3 \\
chaser & 5.6 & 88.3 & 28.6 & ninja & 0.8 & 3.9 & 11.8 \\
climber & 0.2 & 1.0 & -0.1 & plunder & 2.8 & 7.0 & 8.2 \\
coinrun & 4.8 & 108.1 & 11.8 & pong & -35.8 & 105.0 & 102.6 \\
dodgeball & 0.2 & 4.2 & 5.7 & qbert & 36.2 & 289.2 & 320.0 \\
freeway & 0.0 & 0.0 & 8.8 & seaquest & 102.5 & 746.7 & 730.4 \\
frostbite & 32.5 & 255.0 & 253.3 & space\_invaders & 340.0 & 395.4 & 391.7 \\
fruitbot & -2.0 & 6.4 & 13.1 & starpilot & 1.8 & 35.7 & 41.0 \\
\bottomrule
\end{tabular}
\end{table}

\section{Benchmark Details}
\label{app:bench}
\subsection{Setup and baselines}
\label{app:baselines}

There are three settings in which we measure throughput: (1) a single core setting, (2) a multi-thread setting, and a (3) multi-thread setting with a trainer attached.
For all three, we ran comparisons of our 8 ALE clones and 16 ProcGen clones against their originals, with both arms of every comparison run in a single job on one node, resets included in every timed region, and random actions wherever no trainer is attached.

\begin{table}[H]
\centering
\footnotesize
\setlength{\tabcolsep}{0.5em}
\caption{The three throughput measurements and PlayTrain's speedup over each, as a geometric mean over the shared games. Frame skip is 1. Per-core rows compare against ALE and ProcGen as shipped, thread scaling against a tuned EnvPool. Unmatched means ALE emits its native $210{\times}160$ while PlayTrain and ProcGen render $64{\times}64$.}
\begin{tabular*}{\linewidth}{@{\extracolsep{\fill}}llccrr@{}}
\toprule
measurement & hardware & \begin{tabular}[c]{@{}c@{}}matched\\observation\end{tabular} & \begin{tabular}[c]{@{}c@{}}learner\\attached\end{tabular} & \begin{tabular}[c]{@{}r@{}}speedup\\vs.\ ALE\end{tabular} & \begin{tabular}[c]{@{}r@{}}speedup\\vs.\ ProcGen\end{tabular} \\
\midrule
per core (Fig.~\ref{fig:env_efficiency}C, D) & Intel Sapphire Rapids & --- & --- & 12.62$\times$ & 2.18$\times$ \\
thread scaling (Fig.~\ref{fig:env_efficiency}A) & AMD Genoa & \checkmark & --- & 20.80$\times$ & 2.58$\times$ \\
with a trainer (Table~\ref{tab:train-throughput}) & AMD Genoa, 4$\times$H100 & \checkmark & \checkmark & 5.8$\times$ & 2.25$\times$ \\
\bottomrule
\end{tabular*}
\label{tab:bench-setup}
\end{table}

\textbf{Single core.} In the single-core setting, the cost of one step is the quantity of interest, so a fixed number of steps were designated for warmup, and then we divided a fixed number of completed steps by the walltime afterwards.
Our baselines here are simply the ALE via the \texttt{gymnasium/ale-py} package and ProcGen via the \texttt{procgen} package.
For ALE, we used the \texttt{NoFrameskip-v4} prefix games, with \texttt{frameskip} set to 1 with zero action-repeats.
For ProcGen, every game was set to v0, with num\_levels = 0 and start\_level = 0, meaning that we target the full level distribution.
There were seven trials used to benchmark the speeds, measuring 1500 frames total after the first 200 warmup steps were discarded, and we used the mean.
PlayTrain is stepped from C here and the baselines from Python.
Observations are 64$\times$64 RGB for PlayTrain and ProcGen while ALE emits its native 210$\times$160.
PlayTrain therefore has a slight favoring bias in terms of observation resolution.

\textbf{Multi-thread. } For the multi-threaded setting, we build a vecotrized environment, warm it up and then measure speeds over a fixed 12 second window. 
Scaling efficiency is measured by speed divded by the number of threads relative to the lowest thread count (100\% is linear in this case). 
PlayTrain and both baselines' observations are 64$\times$64 RGB with \texttt{frame\_skip} = 1. 
Baselines run on EnvPool 1.2.5 through the Async API with one pool per NUMA domain and with in-pool thread affinitiy enabled. 
We swept multiple batch sizes and selected the fastest, given the aforementioned configuration.
This specific configuration was selected as it is the fastest Atari configuration EnvPool publishes. 

\begin{table}[h]
\centering
\small
\setlength{\tabcolsep}{0.5em}
\renewcommand{\arraystretch}{1.15}
\caption{
The tuned EnvPool configuration behind the thread-scaling comparison. 
}
\begin{tabular}{@{}l>{\raggedright\arraybackslash}p{0.58\linewidth}@{}}
\toprule
parameter & value \\
\midrule
API & \texttt{make\_gymnasium}, driven with \texttt{async\_reset} and \texttt{send}/\texttt{recv} \\
pools per node & one per NUMA domain, each in its own process \\
envs per pool & total $\div$ domains (2{,}048 envs at 80 threads) \\
threads per pool & total $\div$ domains \\
batch size & $\max(16,\; 3 \times \text{threads per pool})$ \\
thread affinity & offset to that domain's first CPU \\
ALE spec & $64{\times}64$ RGB, \texttt{stack\_num=1}, \texttt{frame\_skip=1} \\
ProcGen spec & defaults, already $64{\times}64$ RGB \\
actions & uniform random, sampled per batch \\
\bottomrule
\end{tabular}
\label{tab:envpool-config}
\end{table}

For EnvPool specifically, each NUMA pool got a four-second warmup before the measured window. 
We summed across pools for node speed measurements and then took the geometric average. 

\begin{table}[H]
\centering
\footnotesize
\caption{Thread scaling behind Figure~\ref{fig:env_efficiency}A, geometric mean over the suite of games.}
\begin{tabular}{@{}rrrrrr@{}}
\toprule
 & \multicolumn{2}{c}{env-steps/s} & & \multicolumn{2}{c}{scaling efficiency} \\
\cmidrule(lr){2-3}\cmidrule(lr){5-6}
threads & PlayTrain & EnvPool & ratio & PlayTrain & EnvPool \\
\midrule
\multicolumn{6}{@{}l}{\textit{ProcGen, 16 shared games}} \\
 5 &   227{,}433 &   177{,}338 & 1.28$\times$ & 100\% & 100\% \\
10 &   455{,}799 &   354{,}686 & 1.29$\times$ & 100\% & 100\% \\
20 &   910{,}956 &   584{,}458 & 1.56$\times$ & 100\% & 82\% \\
30 & 1{,}368{,}468 &   777{,}384 & 1.76$\times$ & 100\% & 73\% \\
40 & 1{,}819{,}564 &   931{,}467 & 1.95$\times$ & 100\% & 66\% \\
60 & 2{,}741{,}906 & 1{,}197{,}708 & 2.29$\times$ & 100\% & 56\% \\
80 & 3{,}650{,}005 & 1{,}412{,}903 & \textbf{2.58}$\times$ & 100\% & \textbf{50\%} \\
\midrule
\multicolumn{6}{@{}l}{\textit{ALE, 8 shared games}} \\
 5 &   460{,}983 &    22{,}806 & 20.21$\times$ & 100\% & 100\% \\
10 &   920{,}486 &    45{,}600 & 20.19$\times$ & 100\% & 100\% \\
20 & 1{,}846{,}763 &    89{,}270 & 20.69$\times$ & 100\% & 98\% \\
30 & 2{,}760{,}813 &   134{,}145 & 20.58$\times$ & 100\% & 98\% \\
40 & 3{,}692{,}641 &   177{,}666 & 20.78$\times$ & 100\% & 97\% \\
60 & 5{,}517{,}172 &   265{,}505 & 20.78$\times$ & 100\% & 97\% \\
80 & 7{,}300{,}384 &   350{,}959 & \textbf{20.80}$\times$ &  99\% & \textbf{96\%} \\
\bottomrule
\end{tabular}
\label{tab:bench-scaling}
\end{table}

\textbf{With a trainer.} With a trainer attached, we run four timed windows per game and report the median agent-steps/s. 
We take step counts from the trainer's own counter for measurements since the trainer drives the stepping process.

\subsection{Environment Implementation Details}\label{app:envcost}

\begin{figure}[p]
\centering
\includegraphics[width=\linewidth]{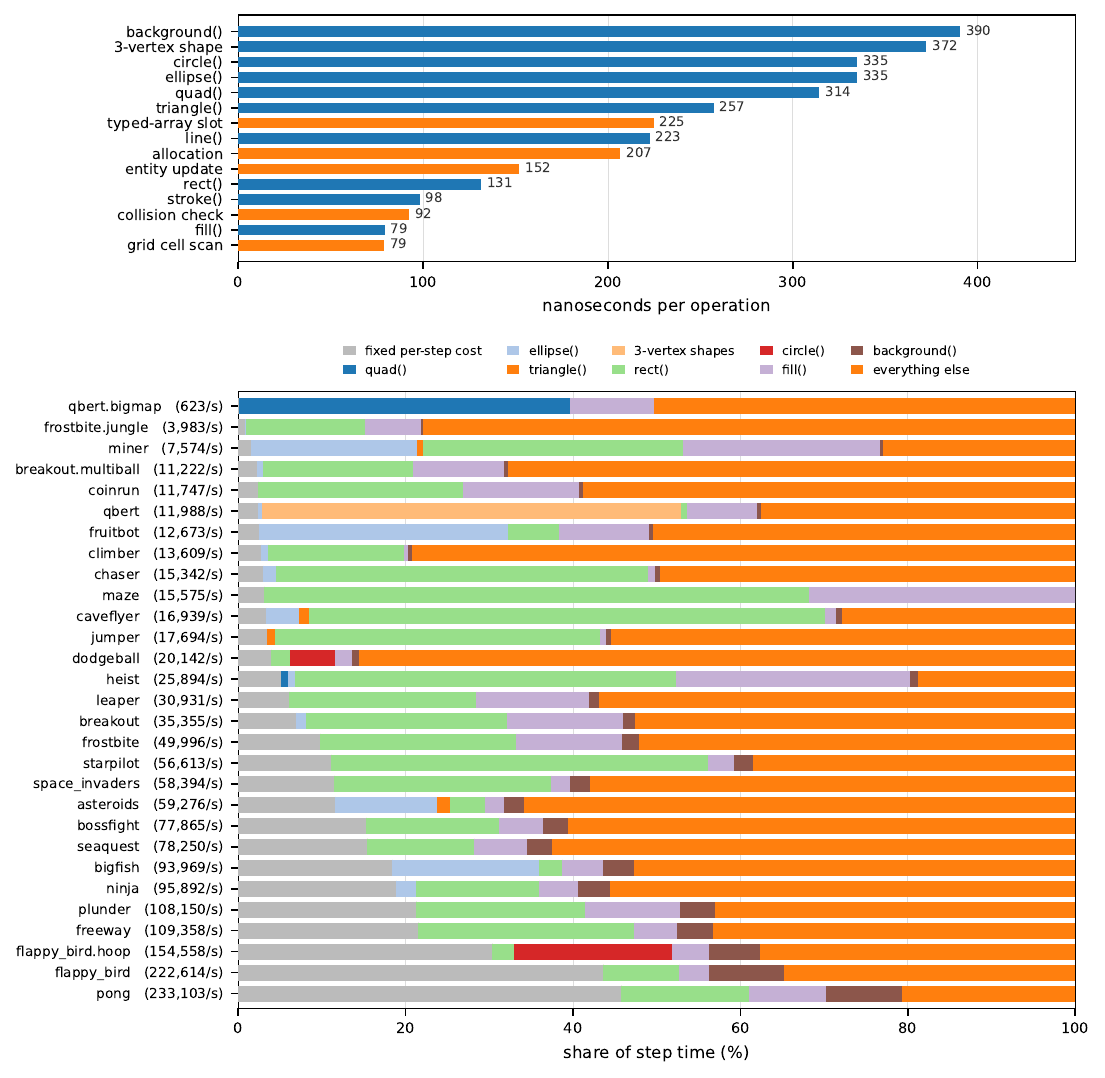}
\caption{\textbf{(A)} Cost of one operation---p5 drawing primitives and game-logic operations---each measured on a probe environment that varies a single quantity. Shapes are priced per polygon, since vertices inside one \texttt{beginShape} accumlates into a single shape. \textbf{(B)} Every game ordered by cost per step with steps per second in brackets, split into the fixed cost. Each primitive's share here is the call count multiplied by the cost in \textbf{(A)}. The remainder is allocated to game logic and interpreter time. Each is measured on one core with dirty-rectangle skipping off. \texttt{maze} is the one game whose cost operations exceed its step time, so its bar doesn't show the residual game logic/interpreter time.}
\label{fig:envcost}
\end{figure}

Here we decompose a step's cost, using probes to measure each operation's latency.
Sec.~\ref{sec:discussion} notes that throughput depends on how many logic and drawing operations a game runs. 
Counting logic and drawing operations by itself does not say which of the two is more costly.
For instance, there are some relatively slower PlayTrain games that barely have any drawing commands while some of the faster games spend 50\% of their \texttt{step()} for drawing. 
This is why we find profiling to be critical because it brings clarity regarding what makes an environment slow or fast \textit{beyond} counting drawing or game logic operations. 

How can we actually profile this problem? 
Drawing and game logic costs are confounded in a real game, so intead we meausre each operation in its own separate environment. 
This takes shape as a generated PlayTrain file whose \texttt{draw()} clears the game canvas and calls N copies of one operation at random positions (N goes up to 4096).
Game logic is measured in the same way but with the drawing pinned instead. 
Now, in order to profile an actual game, we count its calls per operation, multiply each count by that operation's cost from Figure~\ref{fig:envcost}, and then divide the total by the game's step latency (the leftover time is deemed as game logic). 
All runs use a single core of one Sapphire Rapids node without any dirty-rectangle skipping. 

\begin{table}[h]
\centering\small
\setlength{\tabcolsep}{0.4em}
\caption{Number of p5 drawing commands issued, the percentage of the step spent drawing, and single-core environment speeds per game.}
\label{tab:envcost}
\begin{tabular}{@{}lrrr@{\hskip 1.6em}lrrr@{}}
\toprule
game & cmds & draw \% & steps/s & game & cmds & draw \% & steps/s \\
\midrule
\texttt{pong} & 11 & 34 & 233{,}103 & \texttt{heist} & 273 & 76 & 25{,}894 \\
\texttt{flappy\_bird} & 6 & 22 & 222{,}614 & \texttt{dodgeball} & 30 & 10 & 20{,}142 \\
\texttt{flappy\_bird.hoop} & 10 & 32 & 154{,}558 & \texttt{jumper} & 175 & 41 & 17{,}694 \\
\texttt{freeway} & 25 & 35 & 109{,}358 & \texttt{caveflyer} & 298 & 69 & 16{,}939 \\
\texttt{plunder} & 28 & 36 & 108{,}150 & \texttt{maze} & 638 & 105 & 15{,}575 \\
\texttt{ninja} & 20 & 26 & 95{,}892 & \texttt{chaser} & 231 & 47 & 15{,}342 \\
\texttt{bigfish} & 15 & 29 & 93{,}969 & \texttt{climber} & 98 & 18 & 13{,}609 \\
\texttt{seaquest} & 24 & 22 & 78{,}250 & \texttt{fruitbot} & 215 & 47 & 12{,}673 \\
\texttt{bossfight} & 25 & 24 & 77{,}865 & \texttt{qbert} & 600 & 60 & 11{,}988 \\
\texttt{asteroids} & 19 & 23 & 59{,}276 & \texttt{coinrun} & 308 & 39 & 11{,}747 \\
\texttt{space\_invaders} & 40 & 31 & 58{,}394 & \texttt{breakout.multiball} & 246 & 30 & 11{,}222 \\
\texttt{starpilot} & 69 & 50 & 56{,}613 & \texttt{miner} & 787 & 75 & 7{,}574 \\
\texttt{frostbite} & 69 & 38 & 49{,}996 & \texttt{frostbite.jungle} & 487 & 21 & 3{,}983 \\
\texttt{breakout} & 103 & 40 & 35{,}355 & \texttt{qbert.bigmap} & 4{,}048 & 50 & 623 \\
\texttt{leaper} & 111 & 37 & 30{,}931 & & & & \\
\bottomrule
\end{tabular}
\end{table}

\section{Game Generation}
\label{app:interface}
Here we cover the the interface where games are playtested and refined. We also present the prompts that produces and refines the games. 
\begin{figure}[t]
\centering
\includegraphics[width=\linewidth]{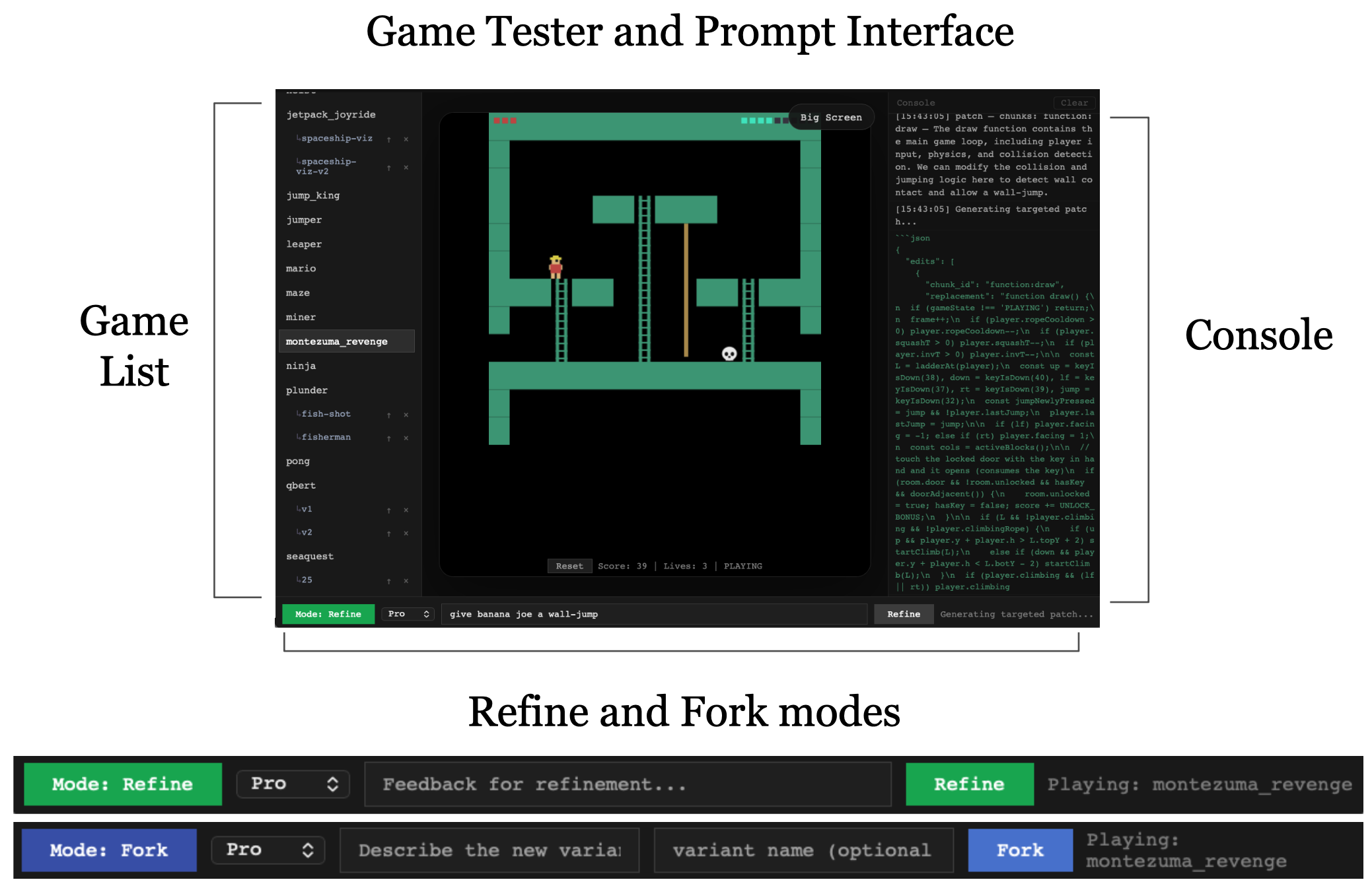}
\caption{The game tester. The left pane lists every game with its variants nested underneath, the center panel runs the selected game in a canvas through the the PlayTrain rasterizer, and the right pane shows the game's console output alongside the model's generation process. The bar underneath switches between Refine and Fork.}
\label{fig:game-interface}
\end{figure}

\begin{figure}[t]
\centering
\includegraphics[width=\linewidth]{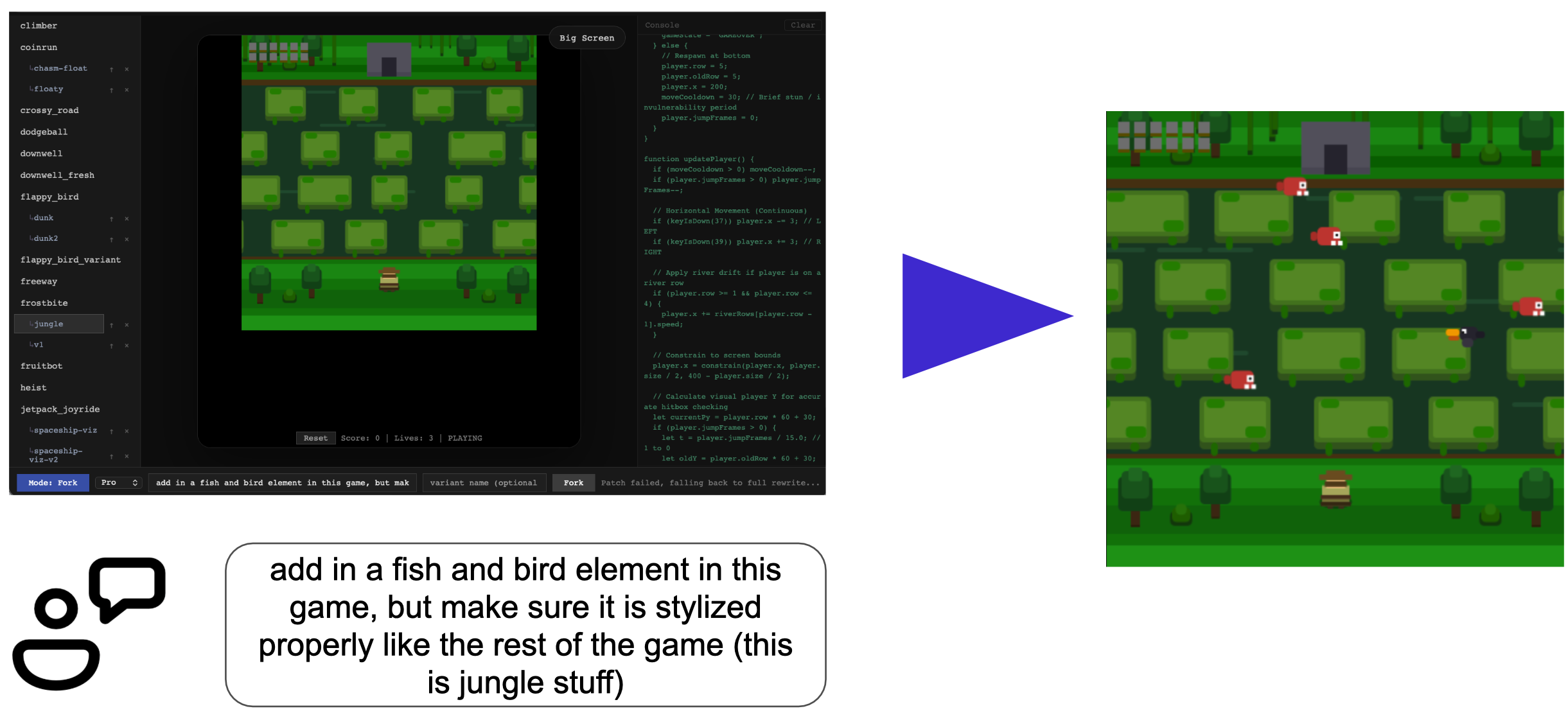}
\caption{A variant can be generated from a single prompt. The text shown is the human's prompt feedback. The left game frame is the original game and the right game frame is the resulting new variant.}
\label{fig:variant-generation-example}
\end{figure}

The game tester renders through our \texttt{raster.mjs} file, which was detailed in Appendix~\ref{app:backend}, while the game itself runs in a canvas at the center of the screen on the browser's animation frame loop.
Forking and refining are the two modes within the prompt interface, but they are essentially the same procedure.
Forking writes a new \texttt{.js} file instead of updating the preexisting one, and asks for a deliberate design change rather than the smallest fix.
Our interface allows us to prompt an LLM (Gemini 3.1 Pro) to generate these forked variations or refinements easily.
The process goes as follows:
the JS game is broken into chunks by top-level function declarations, and as the prompt edit is read by the model, the model is tasked to reason about which chunk to target in reference to the prompt and explain its reasoning.
Afterwards, a second call is made, asking the LLM to output a JSON format to generate targeted edits to the specific chunks that were selected in the first model call. (if the selection or targeted edit call fails for some reason, a full-rewrite is done).
This file (refined or forked) has to define \texttt{setup}, \texttt{draw}, \texttt{getGameState}, \texttt{resetGame} and \texttt{mulberry32} or it gets rejected.
Lastly, everything that is generated is backed up as timestamped snapshots the interface can restore later on.

\subsection{From prompt to game}\label{app:prompt}

\begin{figure}[H]
\centering
\begin{minipage}{0.85\linewidth}
\begin{lstlisting}[style=jsonfig]
{ "name": "donkey_kong",
  "ref": "https://ale.farama.org/environments/donkey_kong/",
  "actions_used": ["LEFT", "RIGHT", "UP", "DOWN", "D", "LEFT+D", "RIGHT+D"],
  "mechanic": "climb ladders + jump barrels to rescue" }
\end{lstlisting}
\end{minipage}
\caption{A catalog entry. The reference URL is fetched and included as text.
\texttt{actions\_used} lists the actions the game uses, and the mechanic is one line describing the game mechanic. }
\label{fig:catalog}
\end{figure}

\begin{table}[t]
\centering
\small
\setlength{\tabcolsep}{0.8em}
\caption{Authorship cost and training throughput per artifact. 
Token counts come from the logged prompts and outputs. 
Costs use Gemini 3.1 Pro rates (\$2 per million input tokens and \$12 per million output). 
SPS is measured on the suite's full-node configuration.}
\begin{tabular}{lrrrrrr}
\toprule
Artifact & Calls & Tokens (in / out) & Time & Cost & LoC $\Delta$ & SPS \\
\midrule
breakout.multiball & 3 & 5{,}292 / 4{,}644 & 2.3 min & \$0.07 & $\sim$140 & 355k \\
qbert.bigmap & 2 & 8{,}459 / 10{,}228 & 5.5 min & \$0.14 & $\sim$290 & 39k \\
flappy\_bird.hoop & 5 & 13{,}204 / 10{,}780 & 7.5 min & \$0.16 & $\sim$80 & 354k \\
frostbite.jungle & 8 & 24{,}932 / 25{,}434 & 10.9 min & \$0.36 & $\sim$300 & 239k \\
vvvvvv & 2 & 4{,}797 / 3{,}283 & 2.3 min & \$0.05 & 258 (new) & 356k \\
downwell & 6 & 14{,}234 / 12{,}837 & 5.9 min & \$0.18 & 419 (new) & 354k \\
\midrule
Total & 26 & 70{,}918 / 67{,}206 & 34.4 min & \$0.95 & --- & --- \\
\bottomrule
\end{tabular}
\label{tab:llm-cost}
\end{table}

\begin{table}[h]
\centering\small
\caption{The four validation checks.}
\label{tab:validation}
\begin{tabular}{@{}l>{\raggedright\arraybackslash}p{0.66\linewidth}@{}}
\toprule
check & requirement \\
\midrule
Gymnasium API & passes \texttt{check\_env} \\
Determinism   & same seed and actions on two instances give identical observations \\
Observation   & \texttt{(64, 64, 3)} \texttt{uint8} in $[0, 255]$; $\geq$ 5 unique pixels;
                changes within 30 steps \\
Reward        & equals the score delta; terminated steps report \texttt{WIN}, \texttt{EXIT} or
                \texttt{GAMEOVER} \\
\bottomrule
\end{tabular}
\end{table}

As stated in Figure~\ref{fig:generation}B, cloning game only using an LLM doesn't require many components. 
For our process within the paper, we have a JSON file that outlines the name of the game we are cloning, a 
online reference link for more details, actions used within the proposed cloned environment, and a small 
concise mechanic describing the games general dynamics. 
The link is fetched and pasted into the prompt as plain text, and our ProcGen clones pass the original C\texttt{++} source in place of a mechanic line.
All of these resources are not required, and more contemporary tools such as Anthropic's Claude Code or OpenAI's Codex can be used instead to build PlayTrain environments. The final ingredients are the prompts, each shown exactly as sent: generate
(Listing~\ref{lst:prompt}), refine (Listing~\ref{lst:refine}), and fork (Listing~\ref{lst:fork}).
Generation is a single call to \texttt{gemini-3.1-pro-preview}.

\begin{lstlisting}[caption={Generation prompt for \texttt{downwell}.},label={lst:prompt},style=prompt]
Generate a p5.js game implementing "downwell".

Mechanic: vertical descent shooter
Actions this game should use: LEFT, RIGHT, D


Reference description of the original game:
Downwell is a curious game about a young person venturing down a well in search of untold treasures with only his Gunboots to protect him.
Reviews "Falling with style." 10/10 - Destructoid "A brilliantly balanced vertical roguelike" 4.5/5 Stars - Pocket Gamer "It's a deep dark well filled with monsters and gems, equal parts platformer and reverse vertical shooter, and falling into its depths is the stuff that 1980s arcade dreams were made of." Hardcore Gamer

The game MUST conform to this template specification exactly:

# Game Template Specification

Standard interface for LLM-generated p5.js games targeting headless RL training.

All games MUST conform to this spec. A single RL agent with a fixed CNN policy trains across all games -- the template guarantees a uniform action space, observation space, and state interface.

## Action Space

**Discrete(8)** -- identical across all games. Actions are abstract -- games interpret them however they want. An agent learns what each action does from pixels and rewards, not from labels.

This follows ProcGen's design: ProcGen uses Discrete(15) with abstract directional + button combinations. Each of its 16 games interprets the same actions differently.

| Index | Name | Keys Held | Key Pressed |
|-------|------|-----------|-------------|
| 0 | NOOP | -- | -- |
| 1 | LEFT | <- | -- |
| 2 | RIGHT | -> | -- |
| 3 | UP | ^ | -- |
| 4 | DOWN | v | -- |
| 5 | D | -- | SPACE |
| 6 | LEFT+D | <- | SPACE |
| 7 | RIGHT+D | -> | SPACE |

**D is a generic action button.** Each game decides what it means:

| Game type | LEFT/RIGHT | UP/DOWN | D |
|-----------|------------|---------|---|
| Platformer | Move | Climb/duck | Jump |
| Shooter | Move | Aim | Fire |
| Angry Birds | Aim angle | Adjust power | Launch |
| Suika | Move drop pos | -- | Drop |
| Snake | Turn left/right | Turn up/down | -- (unused) |
| Breakout | Move paddle | -- | -- (unused) |

Games that don't need all 8 actions simply ignore the extras.

Games read input through `keyIsDown(code)` and the `keyPressed()` callback -- same as standard p5.js. The runtime injects key state before each `draw()` call.

Key codes: LEFT_ARROW=37, UP_ARROW=38, RIGHT_ARROW=39, DOWN_ARROW=40, SPACE=32.

**Runtime action mapping (for reference):**

```javascript
const ACTIONS = [
  { name: 'NOOP',    held: [],   press: null },
  { name: 'LEFT',    held: [37], press: null },
  { name: 'RIGHT',   held: [39], press: null },
  { name: 'UP',      held: [38], press: null },
  { name: 'DOWN',    held: [40], press: null },
  { name: 'D',       held: [],   press: 32 },
  { name: 'LEFT+D',  held: [37], press: 32 },
  { name: 'RIGHT+D', held: [39], press: 32 },
];
```

## Observation Space

- Canvas: any size in-game, downscaled to **64x64 RGB** by the runtime
- **No frame stacking** -- single frame, 3 color channels -> final observation shape: `(64, 64, 3)` uint8
- The game does NOT handle downscaling or color conversion
- This matches ProcGen's observation spec exactly

Recommended canvas size: 256x256 to 512x512. Anything that looks readable at 64x64.

## Required Game Interface

Every game file is a single `.js` file that defines these globals:

```javascript
// ============================================================
// REQUIRED: p5.js lifecycle
// ============================================================

function setup() {
  // Create canvas, initialize constants.
  // Do NOT generate level here -- that happens in resetGame().
  createCanvas(400, 400);
}

function draw() {
  // Main game loop. Called once per tick by the runtime.
  // Read input via keyIsDown(), update state, render frame.
  // For Matter.js games: call Matter.Engine.update(engine, 16.67) here.
}

// ============================================================
// REQUIRED: RL interface
// ============================================================

function getGameState() {
  // Return current game state. Called by the runtime after every draw().
  return {
    score: Number,       // cumulative score (reward = delta per step)
    lives: Number,       // remaining lives; 0 triggers GAMEOVER
    gameState: String,   // one of: 'PLAYING', 'WIN', 'GAMEOVER'
  };
}

function resetGame(seed) {
  // Full reset. Called by the runtime to start a new episode.
  // MUST:
  //   1. Initialize the seeded RNG: rng = mulberry32(seed)
  //   2. Reset score to 0, lives to starting value
  //   3. Set gameState to 'PLAYING'
  //   4. Generate the level procedurally using rng
  //   5. Reset all entity positions, timers, and physics state
  // For Matter.js games: clear and rebuild the Matter.js world here.
}

// ============================================================
// REQUIRED: seeded RNG (copy this verbatim)
// ============================================================

let rng = null;

function mulberry32(seed) {
  let t = seed >>> 0;
  return () => {
    t += 0x6D2B79F5;
    let n = Math.imul(t ^ (t >>> 15), t | 1);
    n ^= n + Math.imul(n ^ (n >>> 7), n | 61);
    return ((n ^ (n >>> 14)) >>> 0) / 4294967296;
  };
}

// Use rng() instead of Math.random() for ALL randomness.
// Example: let x = Math.floor(rng() * width);
```

## What the Seed Controls (game-specific procedural generation)

The seed MUST determine:
- Level layout (terrain, platforms, walls, maze structure)
- Entity spawn positions (enemies, collectibles, obstacles)
- Item/powerup placement and types
- Any randomized parameters (enemy speed, gap sizes, spawn timing)
- Visual variation (color palettes, decorative elements) -- encouraged but optional

The seed MUST NOT affect:
- Core mechanics (gravity, movement speed, rules)
- Action mappings
- Reward structure
- Canvas size

## Reward Design

- `score` starts at 0 on reset
- `score` must increase when the agent does something good (collect item, clear obstacle, kill enemy, progress further)
- `score` may decrease on bad events (lose life, hit obstacle) -- use negative deltas sparingly
- The runtime computes `reward = score_now - score_prev` each step
- Design scores so that a random agent gets near-zero reward and a skilled agent gets high reward

## Terminal Conditions

| gameState | Meaning | When |
|-----------|---------|------|
| `'PLAYING'` | Episode in progress | Default after reset |
| `'WIN'` | Agent completed the objective | Level cleared, goal reached |
| `'GAMEOVER'` | Agent failed | Lives == 0, fatal collision |

The runtime also enforces a `maxSteps` truncation (default 2000). Games do not need to handle this.

## Matter.js Games (Physics)

For games requiring rigid body physics (Angry Birds, Suika, etc.):

```javascript
// Matter.js is available as a global: Matter
// Access via: Matter.Engine, Matter.World, Matter.Bodies, etc.

let engine, world;

function setup() {
  createCanvas(400, 400);
  // Do NOT create the engine here -- do it in resetGame()
}

function resetGame(seed) {
  rng = mulberry32(seed);
  score = 0;
  lives = 3;
  gameState = 'PLAYING';

  // Create fresh physics world each reset
  engine = Matter.Engine.create();
  world = engine.world;
  engine.gravity.y = 1;

  // Add ground, walls, etc.
  let ground = Matter.Bodies.rectangle(200, 390, 400, 20, { isStatic: true });
  Matter.World.add(world, [ground]);

  // Procedurally generate level using rng
  generateLevel(rng);
}

function draw() {
  // Fixed timestep physics update (deterministic)
  Matter.Engine.update(engine, 1000 / 60);

  // Render: read body positions, draw with p5.js
  background(200);
  for (let body of Matter.Composite.allBodies(world)) {
    // ... draw body using rect(), ellipse(), etc.
  }

  // Game logic: check collisions, update score, etc.
}
```

**Determinism guarantee**: Matter.js with fixed timestep + identical initial conditions = identical simulation. All initial conditions come from the seeded RNG, so replays are bit-identical.

## File Structure

```
games/
  flappy.js          # Game source (conforms to this template)
  crossy.js
  angry_birds.js     # Matter.js physics game
  suika.js           # Matter.js physics game
  ...
```

Each file is a self-contained game. No imports, no modules -- all game code in a single file. The runtime provides p5.js globals and (optionally) Matter.js globals before execution.

## Visual Design Rules

The agent sees the game as a **64x64 RGB image**. Every visual decision must serve that constraint. Think ProcGen / Atari 2600, not modern mobile game.

### Color and Contrast
- **Black or dark background** -- maximizes contrast with game elements
- **Use color to encode meaning** -- red = danger/enemies, green = collectibles/safe, blue = player, yellow = coins/points. The agent has full RGB, so color IS information
- **Distinct color per element type**: player, enemies, collectibles, and terrain should each be a different hue. Don't use similar colors for different entity types
- **No gradients, shadows, glow effects, or alpha transparency** -- these become muddy blobs at 64x64
- **Solid fills only** -- `fill()` + `rect()`/`ellipse()`, no complex rendering

### Size and Shape
- **Minimum entity size: 6x6 pixels** on the source canvas (scales to ~1px at 64x64 -- edge of visibility). Prefer 10x10+ for important entities
- **Player should be at least 12x12 pixels** on the source canvas
- **Use distinct shapes per entity type**: player = rectangle, enemies = circles, collectibles = small squares, terrain = large rectangles. Shape + color differentiation helps the CNN
- **No fine detail** -- no 1px lines, no small dots, no intricate patterns

### HUD and Text
- **No text-based HUD** -- text is unreadable at 64x64. The agent cannot read "Score: 150"
- **No title screens, menus, or instructions** -- `resetGame()` goes straight to gameplay
- **No pause screens or cutscenes** -- every frame is gameplay
- If you must show score visually, use a **bar or block indicator** at the screen edge, not text

### What NOT to Render
- Decorative backgrounds (starfields, clouds, grass patterns)
- Particle effects (explosions, sparkles, trails)
- Screen shake or visual transitions
- Drop shadows or outlines on entities
- Antialiased or rounded visual flourishes

### Reference Style
Think: **ProcGen**. Flat colored rectangles, circles, and lines on a dark background. Bright, distinct colors per entity type. Every pixel on screen either means something to gameplay or is background.

## Mechanical Simplicity Rules

Games must be simple enough that an RL agent can learn a basic policy within 1-5 million steps. Complexity kills learning.

### Core Mechanic
- **One core mechanic per game** -- "jump over obstacles", "shoot enemies", "collect items while avoiding hazards". Not all three combined
- **The core mechanic must be exercisable within 10 steps** -- the agent shouldn't need 500 steps of preamble before gameplay starts
- **No multi-phase gameplay** -- no "first collect keys, then unlock doors, then fight boss". One continuous loop
- **No inventory, crafting, or resource management** beyond simple counters (lives, ammo)

### Difficulty and Pacing
- **Immediate reward signal** -- the agent should encounter its first positive reward opportunity within 20-50 steps of random play
- **Frequent scoring opportunities** -- at least one chance to score every 50-100 steps
- **Gradual difficulty** -- early seeds/levels should be easy enough that random agents occasionally score; later seeds should be challenging
- **Death should be possible but not instant** -- give the agent a few lives so it can learn from mistakes within an episode

### What NOT to Include
- Shops, upgrades, or progression systems
- Multiple weapon types or character classes
- Story, dialogue, or narrative elements
- Tutorial sequences
- Complex state machines (charge attacks, combo systems, stance switching)

## Constraints for LLM Generation

When prompting an LLM to generate games:

1. **Single file, no imports** -- all game code in one `.js` file
2. **No DOM access** -- no `document.getElementById`, no CSS, no HTML elements
3. **No async/await** -- `draw()` is synchronous
4. **No images/audio** -- render everything with drawing primitives (rect, ellipse, line, text)
5. **No setTimeout/setInterval** -- the runtime controls frame timing via `tick()`
6. **All randomness via `rng()`** -- never use `Math.random()`
7. **Keyboard input only** -- no mouse, no touch, no gamepad
8. **Score must be meaningful** -- a random-action agent should score near zero; a skilled agent should score high
9. **Episodes must terminate** -- games must reach WIN or GAMEOVER within reasonable play, not run forever
10. **ProcGen-style visuals** -- dark background, distinct colors per entity type, solid shapes, no text HUD, no decorations (see Visual Design Rules above)
11. **One core mechanic** -- simple, learnable, immediate reward (see Mechanical Simplicity Rules above)

## Validation Checklist

A game passes validation if:

- [ ] `resetGame(seed)` runs without error
- [ ] `getGameState()` returns `{ score: Number, lives: Number, gameState: 'PLAYING' }`
- [ ] 200 steps with seed=42 + identical actions produce bit-identical frames (determinism)
- [ ] Frames are non-degenerate (not all one color, multiple unique pixel values)
- [ ] Score changes at least once in 500 random-action steps
- [ ] Game reaches GAMEOVER or WIN within 5000 random-action steps
- [ ] No errors/exceptions during 1000 random-action steps


Output ONLY the JavaScript code. No markdown fences, no explanation.
\end{lstlisting}

\begin{lstlisting}[caption={Refinement prompt for \texttt{downwell}.},label={lst:refine},style=prompt]
You are patching selected chunks in a p5.js game.

Game: downwell_fresh
Player feedback: can you add explicit enemies that are flying horizontally sometimes, and the score should only increase when I kill enemies and get the green orbs, and can you make the character look more akin like a human with a simple walk, jump, and shoot animation cycle, and make sure to make the green orbs sparingly, can you make all of these changes. the avatar can stay blue

Critical requirements:
- Keep the file as plain JavaScript for p5.js. No imports, no modules.
- Preserve these required functions: setup, draw, getGameState, resetGame, mulberry32.
- Keep deterministic reset behavior via rng = mulberry32(seed) in resetGame.
- Keep score, lives, and gameState consistent with getGameState().
- Keep the runtime-compatible control pattern using keyIsDown(...) / keyPressed() as needed.
- Do not change the file into TypeScript or add markdown fences/explanations.
- Return valid JavaScript only when asked for code, and valid JSON only when asked for JSON.


Only modify the chunks provided below. Return JSON only:
{
  "edits": [
    {
      "chunk_id": "function:draw",
      "replacement": "complete replacement code for that chunk only"
    }
  ],
  "notes": "brief summary"
}

Rules for replacements:
- For function chunks, replacement must include the full function definition with the same function name.
- For text chunks, replacement must be the full replacement text for that chunk.
- Do not include untouched chunks.
- Do not include markdown fences in the JSON values.

Editable chunks:

[the 306 lines of selected game source follow here, omitted]
\end{lstlisting}

\begin{lstlisting}[caption={Forking prompt for \texttt{frostbite.jungle}.},label={lst:fork},style=prompt]
Here is the current game code:

[the 377 lines of the parent game's source follow here, omitted]

Requested variant:
now instead of a frostbite, make it jungle themed

Rewrite the full file to implement this variant. You may freely change the mechanics, rules, entities, visuals, and difficulty to realize it -- only the technical contract below (required functions, Discrete(8) controls, seeded determinism, getGameState) must stay intact.

Critical requirements:
- Keep the file as plain JavaScript for p5.js. No imports, no modules.
- Preserve these required functions: setup, draw, getGameState, resetGame, mulberry32.
- Keep deterministic reset behavior via rng = mulberry32(seed) in resetGame.
- Keep score, lives, and gameState consistent with getGameState().
- Keep the runtime-compatible control pattern using keyIsDown(...) / keyPressed() as needed.
- Do not change the file into TypeScript or add markdown fences/explanations.
- Return valid JavaScript only when asked for code, and valid JSON only when asked for JSON.


Output ONLY the complete updated JavaScript code. No markdown fences, no explanation.
\end{lstlisting}




\end{document}